\documentclass[10pt,journal,compsoc]{IEEEtran}
\ifCLASSOPTIONcompsoc
\usepackage[nocompress]{cite}
\else
\usepackage{cite}
\fi
\ifCLASSINFOpdf
 \usepackage[pdftex]{graphicx}
\else
\usepackage[dvips]{graphicx}
\DeclareGraphicsExtensions{.eps}
\fi
\usepackage{amsmath}
\usepackage{amssymb}
\usepackage{amsthm}

\usepackage{array}
\usepackage[linesnumbered,ruled]{algorithm2e}

\theoremstyle{definition}

\let\oldnl\nl
\newcommand{\nonl}{\renewcommand{\nl}{\let\nl\oldnl}}

\renewcommand\thesection{\Alph{section}}

\usepackage{multirow}
\usepackage[font=scriptsize]{subfig}

\usepackage{epstopdf}
\usepackage{subfiles}
\usepackage{makecell}
\usepackage{soul}
\usepackage{color, xcolor}

\usepackage{pifont}

\usepackage{verbatim}

\soulregister\cite7 
\soulregister\eqref7
\usepackage{booktabs}
\usepackage{colortbl}
\usepackage{url}

\begin{document}
	%
	\title{Correctable Landmark Discovery via Large Models for Vision-Language Navigation}
	%
	%
	%
	%
	\author{Bingqian~Lin\IEEEauthorrefmark{1}, Yunshuang Nie\IEEEauthorrefmark{1}, Ziming Wei, Yi Zhu, Hang Xu, \\ Shikui Ma, Jianzhuang Liu, Xiaodan Liang\IEEEauthorrefmark{2}
	\IEEEcompsocitemizethanks{
	\IEEEcompsocthanksitem 
	\IEEEauthorrefmark{1}These two authors contribute equally to this work.\protect\\
	\IEEEcompsocthanksitem 
	\IEEEauthorrefmark{2}Xiaodan Liang is the corresponding author.\protect\\	\IEEEcompsocthanksitem Bingqian Lin, Yunshuang Nie and Ziming Wei are with Shenzhen Campus of Sun Yat-sen University, Shenzhen, China. \protect\\
			E-mail:\{linbq6@mail2.sysu.edu.cn,nieysh@mail2.sysu.edu.cn, weizm3@mail2.sysu.edu.cn\}
		\IEEEcompsocthanksitem Xiaodan Liang is with Shenzhen Campus of Sun Yat-sen University, Shenzhen, China, and also with PengCheng Laboratory.
  \protect\\
  E-mail: liangxd9@mail.sysu.edu.cn
		\IEEEcompsocthanksitem Yi Zhu and Hang Xu are with Huawei Noah's Ark Lab. \protect\\
 E-mail: zhu.yee@outlook.com, chromexbjxh@gmail.com. 
 \IEEEcompsocthanksitem Shikui Ma is with Dataa Robotics company. \protect\\
 E-mail: maskey.bj@gmail.com.
 \IEEEcompsocthanksitem Jianzhuang Liu is with Shenzhen Institute of Advanced Technology, Shenzhen, China. \protect\\
 E-mail: jz.liu@siat.ac.cn.
	
	}
		}
	
	%
	%
	
	\markboth{IEEE TRANSACTIONS ON PATTERN ANALYSIS AND MACHINE INTELLIGENCE}%
	{Shell \MakeLowercase{\textit{et al.}}: Bare Demo of IEEEtran.cls for Computer Society Journals}
	%



	
		\IEEEtitleabstractindextext{%
		\begin{abstract}
Vision-Language Navigation (VLN) requires the agent to follow language instructions to reach a target position. 
A key factor for successful navigation is to align the landmarks implied in the instruction with diverse visual observations.
However, previous VLN agents fail to perform accurate modality alignment especially in unexplored scenes, since they learn from limited navigation data and lack sufficient open-world alignment knowledge. 
In this work,  we propose a new VLN paradigm, called \textbf{CO}rrectable La\textbf{N}dmark Di\textbf{S}c\textbf{O}very via \textbf{L}arge Mod\textbf{E}ls (CONSOLE). 
In CONSOLE, we cast VLN as an open-world sequential landmark discovery problem, by introducing a novel correctable landmark discovery scheme based on two large models ChatGPT and CLIP.
Specifically, we use ChatGPT to provide rich open-world landmark cooccurrence commonsense, and
conduct CLIP-driven landmark discovery based on these commonsense priors. 
To mitigate the noise in the priors due to the lack of visual constraints, we introduce a learnable cooccurrence scoring module, which corrects the importance of each cooccurrence according to actual observations for accurate landmark discovery. 
We further design an  observation enhancement strategy for an elegant combination of our framework with different VLN agents, where we utilize the corrected landmark features to obtain enhanced observation features for action decision. 
Extensive experimental results on multiple popular VLN benchmarks (R2R, REVERIE, R4R, RxR) show the significant superiority of CONSOLE over strong baselines. Especially, our CONSOLE establishes the new state-of-the-art results on R2R and R4R in unseen scenarios. Code is available at \url{https://github.com/expectorlin/CONSOLE}.

		\end{abstract}
		
		}

	\maketitle

	\IEEEdisplaynontitleabstractindextext

	%
	\IEEEpeerreviewmaketitle

 \IEEEraisesectionheading{\section{Introduction}\label{sec:introduction}}

	%
	%
	%
	%


\IEEEPARstart{V}{ision-and-Language} Navigation (VLN)~\cite{anderson2018vision,qi2020reverie,jain2019stay,ku2020room,Chen2019TOUCHDOWNNL}, one of the most representative embodied AI tasks, requires an agent to navigate through complicated visual environments to a goal position following a given natural language instruction. It has led to a wide range of research recently since an
instruction-following navigation agent is more flexible and
practical in real-world applications. For successful navigation,  the agent needs to accurately understand the instruction intention and align the landmarks implied in the instruction  to sequential visual observations.

Early VLN methods utilize different data augmentation strategies~\cite{tan2019learning, fried2018speaker,liu2021vision}, efficient learning paradigms~\cite{zhu2020vision,li2019robust,huang2019transferable}, and dedicated model architectures~\cite{ma2019self, hong2020language,deng2020evolving} to learn useful modality alignment knowledge from limited human annotated navigation data. With the development of cross-modal pretraining models, increasing pretraining-based VLN approaches are developed~\cite{hong2021vln,Chen2021HistoryAM} and have shown great progress in improving the modality alignment ability of VLN agents. 
However, due to the limited scale and diversity of pretraining and navigational data, these approaches still cannot generalize well to unseen navigation scenarios, which usually  require rich open-world alignment knowledge, e.g., recognizing the unseen landmark, to realize accurate instruction following.

\begin{figure*}[t]
\begin{centering}
\includegraphics[width=0.98\linewidth]{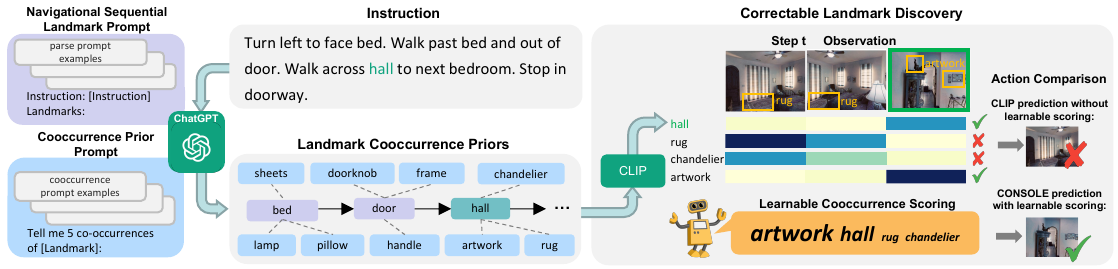}
\par\end{centering}
\caption{Correctable landmark discovery of CONSOLE. We provide customized prompts
for ChatGPT to generate landmark cooccurrence priors. 
Then we introduce a learnable cooccurrence scoring module to conduct CLIP-driven correctable landmark discovery based on the priors. The landmark/cooccurrence with bigger font size has a higher score.
}
\label{fig:motivation}
\vspace{-0.4cm}
\end{figure*}

Recent studies have shown that employing large models, such as Large Language Models (LLMs, e.g., GPT-3~\cite{brown2020language})
and large Vision-Language Models (VLMs, e.g., CLIP~\cite{radford2021learning}), for embodied AI tasks is a very promising way to improve the task completion~\cite{Ahn2022DoAI,Huang2022InnerME,Liang2022VisualLanguageNP}, since the large models are pretrained on ultra-large-scale web corpora and capture vast amount of open-world knowledge which is helpful for embodied AI tasks. However, constrained by the large domain gap between those training corpora and embodied AI task datasets, a direct introduction of large models for embodied AI tasks may bring unexpected noise~\cite{Ahn2022DoAI,Song2022LLMPlannerFG}. Therefore, how to effectively activate and more importantly, utilize the task-related knowledge from large models to assist embodied agents, has been paid more and more attention in the embodied AI area.

In this paper, we propose a new paradigm called \textbf{CO}rrectable La\textbf{N}dmark Di\textbf{S}c\textbf{O}very via \textbf{L}arge Mod\textbf{E}ls (CONSOLE), where we activate and utilize rich open-world alignment knowledge stored in two large models ChatGPT~\cite{openai} and CLIP~\cite{radford2021learning} for assisting the VLN task in a correctable way. 
Through CONSOLE, we cast VLN as an open-world sequential landmark discovery problem for facilitating action decision. 
Our approach is inspired by the navigation commonsense, that is, the landmark cooccurrence is helpful for locating a target landmark. 
For example, a {\it fridge} is likely to be placed near a {\it stove}. 
When a mentioned landmark in the instruction is hardly observable in the views, locating the landmark cooccurrence (which is not mentioned in the instruction) is an effective alternative of locating the mentioned landmark to make correct navigation decisions. 
Being trained using ultra-large-scale corpora, ChatGPT captures sufficient open-world landmark cooccurrence commonsense.
However, due to the lack of visual constraints, these commonsense priors may be inconsistent with actual navigation scenes and mislead the agent to making wrong action decision. 
To mitigate the impact of the noisy priors, we introduce a CLIP-driven correctable landmark discovery scheme, where the priors are corrected in a learnable way according to actual observations. A consistency loss is introduced to constrain the prior correction using the navigation supervision. As a result, at each navigation timestep, the importance of each cooccurrence is re-ranked based on the current observation to suppress the noisy cooccurrences for accurate navigation decision.
As shown in Fig.~\ref{fig:motivation}, through correctable landmark discovery, CONSOLE successfully highlights the important landmark and cooccurrence (hall and artwork) for making correct action decision while getting rid of the impact of the noisy ones (rug and chandelier). 

Our approach contains three core components: 1) \textbf{Landmark cooccurrence prior generation}: We design customized prompts for ChatGPT to extract landmark cooccurrence priors, including navigational sequential landmarks and open-world landmark cooccurrences. 2) \textbf{Correctable landmark discovery}: 
We introduce a learnable cooccurrence scoring module constrained by a consistency loss for accurate landmark discovery based on the priors. The scoring module ranks the importance of each cooccurrence according to dynamic navigation observations. 
3) \textbf{Observation enhancement}: We use corrected landmark features for enhancing observation features to combine the landmark discovery framework with different VLN agents. The enhanced observation features are adopted for final action decision.

We evaluate CONSOLE on multiple mainstream VLN benchmarks, including R2R~\cite{anderson2018vision}, REVERIE~\cite{qi2020reverie},  R4R~\cite{jain2019stay}, and RxR~\cite{ku2020room}. 
The experimental results show that CONSOLE outperforms strong baselines~\cite{Chen2021HistoryAM, Chen2022ThinkGA} on all benchmarks. Moreover, CONSOLE achieves the new state-of-the-art on R2R and R4R in unseen scenarios. We conduct extensive insightful visualization to verify the effectiveness and necessity of different components in CONSOLE, such as the learnable scoring module and the customized designs for the prompts.  
The visualization comprehensively reveals how CONSOLE successfully activates and utilizes the helpful navigation knowledge in large models for assisting action decision. We also present both quantitative and qualitative comparison of different LLMs to deeply analyze their potential in assisting VLN tasks. 
We believe that our work will provide a non-trivial reference in introducing large models for assisting embodied AI tasks in a noise-suppressed way.

To summarize, our main contributions are:
\begin{itemize}
\item{We propose a novel VLN paradigm called CONSOLE, where we activate and utilize the rich open-world knowledge in large models in a correctable way, which effectively assists VLN while mitigating the misleading brought by large models.}

\item{We cast VLN as an open-world sequential landmark discovery problem to effectively harvest the knowledge helpful for VLN from both LLM and VLM. Built upon an observation enhancement strategy, our correctable landmark discovery scheme can be combined with different VLN models elegantly.}

\item{CONSOLE outperforms strong baselines on multiple popular VLN benchmarks, including R2R, REVERIE, R4R and RxR. It also establishes new SOTA results on R2R and R4R in unseen scenarios.}

\end{itemize}

 \section{Related Work}

\subsection{Vision-Language Navigation.}
In Vision-Language Navigation (VLN), the agent needs to move across complicated visual environments to reach a goal location following human instructions~\cite{anderson2018vision,ku2020room,qi2020reverie}. 
Therefore, accurate modality alignment is crucial for successful navigation.
Early methods mainly design diverse data augmentation techniques~\cite{tan2019learning, fried2018speaker,liu2021vision,Fu2019CounterfactualVN}, efficient learning mechanisms~\cite{wang2019reinforced,zhu2020vision,huang2019transferable, li2019robust,Wang2020ActiveVI,Ma2019TheRA,Ke2019TacticalRS}, and useful model architectures~\cite{wang2019reinforced,ma2019self, deng2020evolving,qi2020Object} for learning useful modality alignment knowledge.
EnvDropout~\cite{tan2019learning} creates the augmentation data by mimicking unseen scenes via an environmental dropout scheme. AuxRN~\cite{zhu2020vision} designs multiple self-supervised auxiliary training tasks to exploit rich semantic information in the environment. OAAM~\cite{qi2020Object} designs a new model structure and loss for separately processing the mentioned objects and actions in the instructions.

Recent VLN works employ cross-modal pretraining paradigms and models for enhancing the modality alignment~\cite{hong2021vln,Chen2021HistoryAM,Qi2021TheRT,Chen2022ThinkGA,Zhao2022TargetDrivenST,Qiao2022HOPHA,Guhur2021AirbertIP}. 
HAMT~\cite{Chen2021HistoryAM} introduces a history-aware multimodal transformer to encode the long-horizon navigation history. DUET~\cite{Chen2022ThinkGA} builds a dual-scale graph transformer to jointly perform long-term action planning and fine-grained cross-modal understanding. HOP~\cite{Qiao2022HOPHA} introduces a new history-and-order aware pretraining paradigm for encouraging the learning of spatio-temporal multimodal correspondence.

In this paper, we resort to existing powerful large models to cast VLN as an open-world sequential landmark discovery problem, for facilitating action decision in unseen scenes.
we introduce a novel correctable landmark discovery scheme to correct the priors provided by large models according to actual  observations in a learnable way. 
By using corrected landmark features for enhancing the observation features, we fulfill an elegant combination of our framework with different VLN agents. 

Although some VLN approaches have also explored to decouple the landmark discovery for better cross-modality alignment~\cite{qi2020Object,Moudgil2021SOATAS,Qi2021TheRT}, they tend to use the limited information in the instruction for aligning with the visual observation. In contrast, our method uses large models to provide rich commonsense prior accompanied by a powerful learnable prior correction paradigm, which enables the agent to learn much more open-world alignment knowledge for successful navigation.  

\subsection{LLMs \& VLMs for Embodied Tasks.}
Recent studies have shown that LLMs and VLMs can connect human instructions to robot actions and facilitate the alignment between instructions and scenes~\cite{Huang2022LanguageMA,Ahn2022DoAI,Huang2022InnerME,shahlm,Chen2022OpenvocabularyQS,Liu2023API,Song2022LLMPlannerFG,Dorbala2023CanAE,Lu2022NeuroSymbolicPP, Gadre2022CoWsOP}, thanks to rich real-world and open-world knowledge stored in them. Planner~\cite{Huang2022LanguageMA} employs an LLM to decompose the abstract task instructions to concrete step-by-step executable actions for embodied agents. SayCan~\cite{Ahn2022DoAI} further uses an LLM for assisting the action decision by grounding the LLM through value functions of pretrained robotic skills. 
Inner Monologue~\cite{Huang2022InnerME} introduces a VLM to provide environment feedbacks for an LLM to enable it to recover from failure in robotic control scenes.
LM-Nav~\cite{shahlm} combines an LLM, a VLM, and a visual navigation model to enable long-horizon instruction following
without requiring user-annotated navigational data. 
However, directly using an LLM and/or a VLM for assisting VLN may introduce unexpected noise that confuses the action decision, due to the limited action space of VLN and the large domain gap between the training corpora of large models and the VLN dataset. 

In this work, we activate open-world alignment knowledge from both LLM and VLM to realize an open-world landmark discovery paradigm for assisting VLN. To the best of our knowledge, we are the first to harvest landmark cooccurrence knowledge from LLM to assist the VLN task. Through a correctable landmark discovery scheme, the noise in the priors can be effectively suppressed for accurate action decision. 

A concurrent work ESC~\cite{Zhou2023ESCEW}
also introduces the landmark coocurrence knowledge from LLMs for assisting the zero-shot object navigation task. However, our CONSOLE is different from it in the following two key aspects: 1) we introduce a learnable cooccurrence scoring module to effectively mitigate the noise in the knowledge, which does not require complicated exploration procedure like Zhou et al.~\cite{Zhou2023ESCEW} and hardly impacts the inference speed. This module can also be easily optimized through navigation supervision. 2) Different from Zhou et al.~\cite{Zhou2023ESCEW} that focus on the zero-shot setting, our work proposes a trainable scheme to elegantly combine LLM, VLM, and the VLN agent, which achieves significantly  higher navigation performance than the zero-shot setting.


\section{Preliminaries}
\label{Preliminaries}
\subsection{Problem Setup}
In the discrete VLN task, an agent is initialized at a starting node and needs to explore the navigation connectivity graph $\mathcal{G}=(V,E)$ to reach a target node, following a language instruction $I$. $V$ and $E$ represent the nodes and edges in the navigation connectivity graph. At each timestep $t$, the agent receives a panoramic observation $O_{t}$ of its current node. $O_{t}$ composed of $N_{o}$ single-view observations $O_{t,n}$, i.e., $O_{t}=\{O_{t,n}\}_{n=1}^{N_{o}}$. Each $O_{t,n}$ contains an RGB image $B_{t,n}$ and the directional information. With the instructions $I$ and current visual observations $O_{t}$, the agent infers the action $\mathbf{a}_{t}$ for each step $t$ from the candidate navigable views  $\mathcal{N}(O_{t})\in O_{t}$, which corresponds to  the neighbor nodes of the current node.

In this paper, we propose to cast VLN as an open-world landmark discovery problem to introduce large models for assisting VLN tasks. We define the problem in the following. Specifically, the goal of the open-world sequential landmark discovery is to discover the mentioned landmarks that the agent may not see before sequentially to complete the navigation. Denote the landmark list extracted from the instruction $I$ as $U^{la}$. At each timestep $t$, the agent first needs to predict the landmark $U^{la}_{t}$ which is needed to find currently from the landmark list $U^{la}$. Then, it discovers $U^{la}_{t}$ in the set of navigable viewpoints and makes the action prediction $\mathbf{a}_{t}$ by selecting the navigable viewpoint containing $U^{la}_{t}$.

\subsection{Large Models}
Benefiting from ultra-large-scale web corpora, large pretrained models have shown excellent generalization abilities to various tasks due to their vast knowledge storage. 
In CONSOLE, we resort to two powerful large models ChatGPT~\cite{openai} and CLIP~\cite{radford2021learning}, to provide rich open-world alignment knowledge for facilitating action decision.

ChatGPT~\cite{openai} is a powerful text generation model that shows great in-context reasoning ability, i.e., given with customized prompts with a few task examples, its internalized knowledge about some specific task can be activated through text completion. 

CLIP~\cite{radford2021learning} is a vision-language model pretrained on 400M image-text pairs collected from the web. It utilizes a ViT~\cite{dosovitskiy2021an} or a ResNet-50~\cite{he2016deep} as the image encoder and a Transformer~\cite{vaswani2017attention} as the text encoder. 
By conducting a simple similarity calculation between images and textual class descriptions, 
it shows powerful zero-shot object recognition ability in diverse datasets~\cite{radford2021learning}.

\subsection{Baseline VLN Agents}
We choose two strong baselines HAMT~\cite{Chen2021HistoryAM} and DUET~\cite{Chen2022ThinkGA} to verify the effectiveness of CONSOLE. In this section, we briefly describe one baseline HAMT. In HAMT, the agent receives the instruction $I$, the observation $O_{t}$, and the navigation history $H_{t}$ at each timestep $t$.  
For simplicity, we omit the encoding process for the navigation history $H_{t}$.
Specifically, a text encoder $E^{I}(\cdot)$ and a vision encoder $E^{V}(\cdot)$ are employed to obtain the instruction feature $\mathbf{f}_{I}$ and the observation feature $\mathbf{f}_{O_{t}}$, respectively:
\begin{equation}
\label{unimodal features}
\mathbf{f}_{I}=E^{I}(I), \quad\mathbf{f}_{O_{t}}=E^{V}(O_{t}).
\end{equation}
Then $\mathbf{f}_{I}$ and $\mathbf{f}_{O_{t}}$ are updated through a cross-modal Transformer encoder $E^{c}(\cdot)$: 
\begin{equation}
\label{eq:ec}
\mathbf{\tilde{f}}_{I}^{t},\mathbf{\tilde{f}}_{O_{t}}=E^{c}(\mathbf{f}_{I},\mathbf{f}_{O_{t}}).
\end{equation}
The action prediction probability $\mathbf{a}_{t}$ is generated through an action prediction module $E^{a}(\cdot)$ based on  $\mathbf{\tilde{f}}_{I}^{t}$ and  $\mathbf{\tilde{f}}_{O_{t}}$:
\begin{equation}
\label{eq:ea}
\mathbf{a}_{t} = E^{a}(\mathbf{\tilde{f}}_{I}^{t}, \mathbf{\tilde{f}}_{O_{t}}).
\end{equation}
The action decision is optimized by the  navigation loss $\mathcal{L}_{\mathrm{nav}}$, which contains an imitation learning loss $\mathcal{L}_{\mathrm{IL}}$ and a reinforcement learning loss $\mathcal{L}_{\mathrm{RL}}$~\cite{Chen2021HistoryAM} :
\begin{equation}
\mathcal{L}_{\mathrm{IL}}=\sum_{t}-\mathbf{a}_{t}^{*}\mathrm{log}(\mathbf{a}_{t}),
\end{equation}
\begin{equation}
\mathcal{L}_{\mathrm{RL}}=\sum_{t}-\mathbf{a}_{t}^{s}\mathrm{log}(\mathbf{a}_{t})A_{t},
\end{equation}
\begin{equation}
\mathcal{L}_{\mathrm{nav}} = \mathcal{L}_{\mathrm{IL}}+\lambda\mathcal{L}_{\mathrm{RL}},
\end{equation}
where $\mathbf{a}_{t}^{*}$ is the teacher action of the ground-truth path at timestep $t$, $\mathbf{a}_{t}^{s}$ is the sampled action from the agent action prediction $\mathbf{a}_{t}$, $A_{t}$ is the advantage calculated by A2C algorithm~\cite{Mnih2016AsynchronousMF}, and $\lambda$ is a balance factor.

In CONSOLE, we use the corrected landmark features $\mathbf{f}_{U_{t}}$ and the observation features $\mathbf{f}_{O_{t}}$  to obtain the enhanced observation features $\mathbf{f'}_{O_{t}}$ at each timestep $t$. Then $\mathbf{f'}_{O_{t}}$ are fed into $E^{c}(\cdot)$ and $E^{a}(\cdot)$ to get final action prediction $\mathbf{\tilde{a}}_{t}$.

\begin{table}
\centering
	\fontsize{5}{5}\selectfont
	\caption{Notation summarization of variables. 
 \vspace{-0.2cm}
	}\label{tab:notations}
		\resizebox{0.8\linewidth}{!}{
	{\renewcommand{\arraystretch}{1.1}	\begin{tabular}{ll}
				\specialrule{.1em}{.05em}{.05em}	
			
	Notations&Variables\\ \hline
	
	$I$&instruction\\
    $O_{t}$&observation\\
    $U$&landmark cooccurrence prior\\
    $z$&shifting pointer\\
    $U^{la}_{t}$&current important landmark\\
    $s^{la}$&landmark score\\
    $\{s_{i}^{co}\}_{i=1}^{N^{co}}$&cooccurrence score\\
    $\mathbf{f}_{I}$&instruction feature\\
    $\mathbf{f}_{O_{t}}$&observation feature\\
    $\mathbf{f'}_{O_{t}}$&enhanced observation feature\\
    $\mathbf{f}_{S_{t}}$&state feature\\
    $\mathbf{f}_{U_{t}}$&corrected landmark feature\\
    $\mathcal{L}_{\mathrm{nav}}$&navigation loss\\
    $\mathcal{L}_{\mathrm{cs}}$&consistency loss\\
    $\mathcal{L}_{\mathrm{ct}}$&contrastive loss\\

	\specialrule{.1em}{.05em}{.05em}	\end{tabular}}}
\vspace{-0.2cm}	
\end{table}

\begin{figure*}[t]
\begin{centering}
\includegraphics[width=0.9\linewidth]{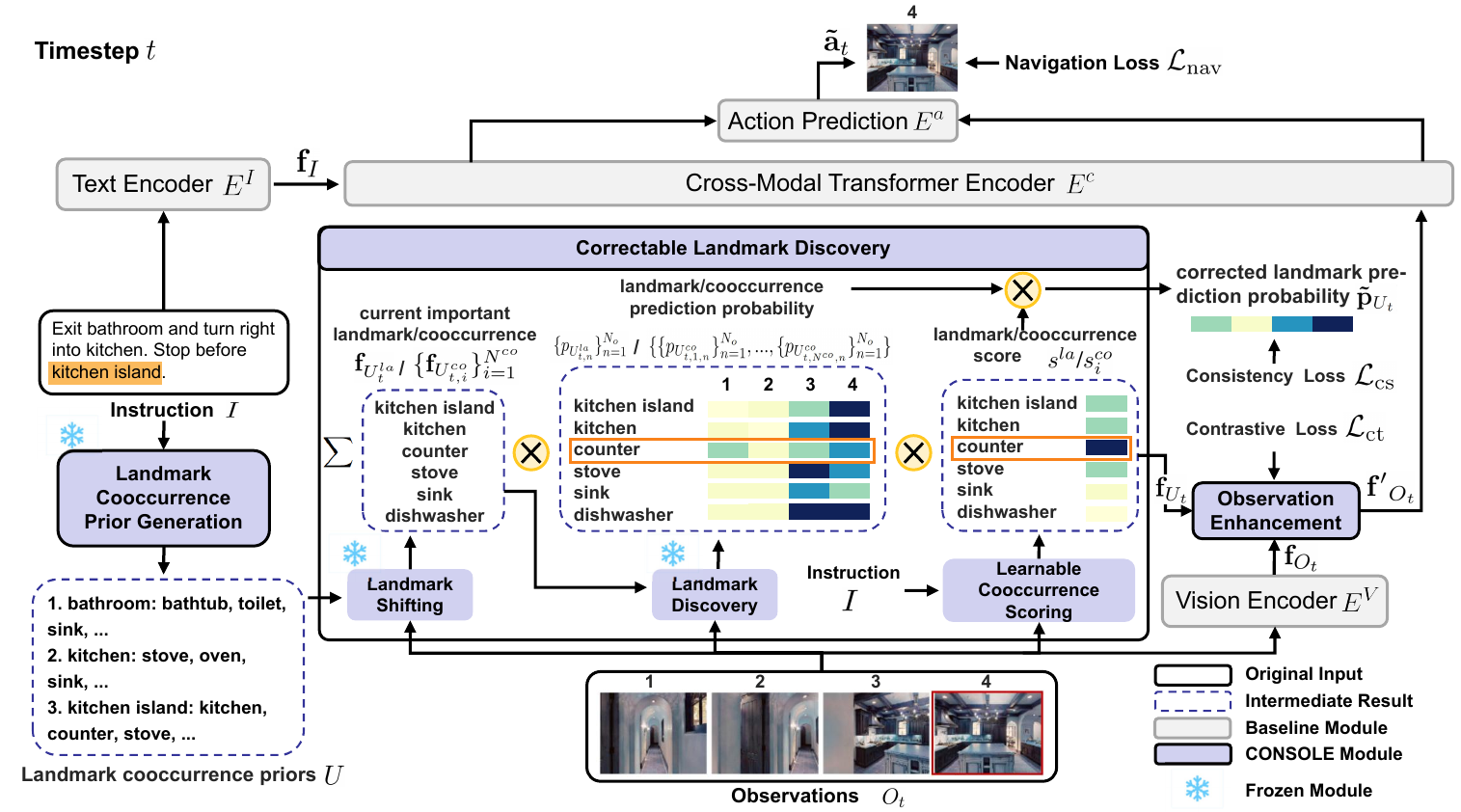}
\par\end{centering}
\caption{Overview of CONSOLE. Before navigation, the landmark cooccurrence priors $U$ are obtained through the landmark cooccurrence prior generation module. At navigation timestep $t$, the agent conducts the correctable landmark discovery based on the landmark shifting, the landmark discovery, and the learnable cooccurrence scoring. An observation enhancing module is introduced to enhance the observation features $\mathbf{f}_{O_{t}}$ using the corrected landmark features $\mathbf{f}_{U_{t}}$. And the enhanced observations $\mathbf{f'}_{O_{t}}$ are used for action decision. Besides the navigation loss $\mathcal{L}_{\mathrm{nav}}$, we introduce the consistency loss $\mathcal{L}_{\mathrm{cs}}$ and the contrastive loss $\mathcal{L}_{\mathrm{ct}}$ for optimization. 
}
\label{fig:overview}
\vspace{-0.4cm}
\end{figure*}

\section{Method}
The overview of CONSOLE is presented in Fig.~\ref{fig:overview}. For a given instruction $I$, we first conduct the  landmark cooccurrence prior generation to obtain the landmark cooccurrence priors $U$ (Sec.~\ref{Landmark Cooccurrence Prior Generation}). 
Then, at each timestep $t$, we perform the correctable landmark discovery through the landmark shifting, the landmark discovery, and the learnable cooccurrence scoring, based on the priors $U$, the instruction $I$, and the observations $O_{t}$  (Sec.~\ref{Correctable Landmark Discovery}).
After the
correctable landmark discovery, we obtain the corrected landmark feature $\mathbf{f}_{U_{t}}$and use it to enhance the observation feature $\mathbf{f}_{O_{t}}$ (Sec.~\ref{Observation Enhancing}). 
We adopt the enhanced observation feature $\mathbf{f'}_{O_{t}}$ for predicting the final action $\mathbf{\tilde{a}}_{t}$.
The total training objective of CONSOLE contains the navigation loss $\mathcal{L}_{\mathrm{nav}}$ and two additional introduced losses, i.e., a  consistency loss $\mathcal{L}_{\mathrm{cs}}$ and a contrastive loss $\mathcal{L}_{\mathrm{ct}}$ (Sec.~\ref{Action Prediction}).
For facilitating reading, we list the crucial variables in CONSOLE in Table~\ref{tab:notations}.

\subsection{Landmark Cooccurrence Prior Generation}
\label{Landmark Cooccurrence Prior Generation}
The landmark cooccurrence prior contains the following two elements: 1) Navigational Sequential Landmarks: 
Due to the complexity of natural language, the order of landmarks appearing in the instruction may be sometimes inconsistent with the intended order to be searched during navigation.
Therefore, extracting navigational sequential landmarks is crucial for the landmark discovery. 2) Landmark Cooccurrence: the open-world landmark cooccurrence knowledge, e.g., a {\it counter} may appear in a {\it kitchen island} (Fig.~\ref{fig:overview}), can greatly help the alignment. 
We design customized prompts with heuristic task examples to obtain these two kinds of priors.

\noindent\textbf{Navigational Sequential Landmark Extraction.} 
An ideal extraction of landmarks requires: (1) extraction of all mentioned landmarks; (2) successful extraction of landmarks with complex descriptions (e.g., landmarks with adjectives or phrases); (3) avoidance of abstract nouns (e.g., right); (4) navigational order instead of textual order.
To ensure the coverage for diverse descriptions of landmarks for accurate extraction, we pre-define 5 different task examples for ChatGPT.
Taking REVERIE as an example, one task example is as follows:

\texttt{Instruction:  Go to the lounge on the first level and bring the trinket that's sitting on the fireplace.}

\texttt{Landmarks: }

\texttt{1.first level; }

\texttt{2.lounge; }

\texttt{3.fireplace; }

\texttt{4.trinket. }

\noindent 
Through such example, we can let ChatGPT know that it should extract navigational sequential landmarks rather than textual sequential landmarks (1.lounge; 2.first level; 3.trinket; 4.fireplace). 
Moreover, it should extract ``first level'' as the landmark rather than ``level''. 
After in-context learning with task examples, we can ask ChatGPT to extract the landmarks of a specific instruction $I$ with the following prompt:

\texttt{Instruction: $I$}
          
\texttt{Landmarks:}

\noindent which is shown in Fig.~\ref{fig:motivation} (upper left). We constrain ChatGPT not to generate abstract landmarks and landmarks not mentioned in the instruction by stating it explicitly in the prompt. 

\begin{figure}
  \centering
   \includegraphics[width=0.9\linewidth]{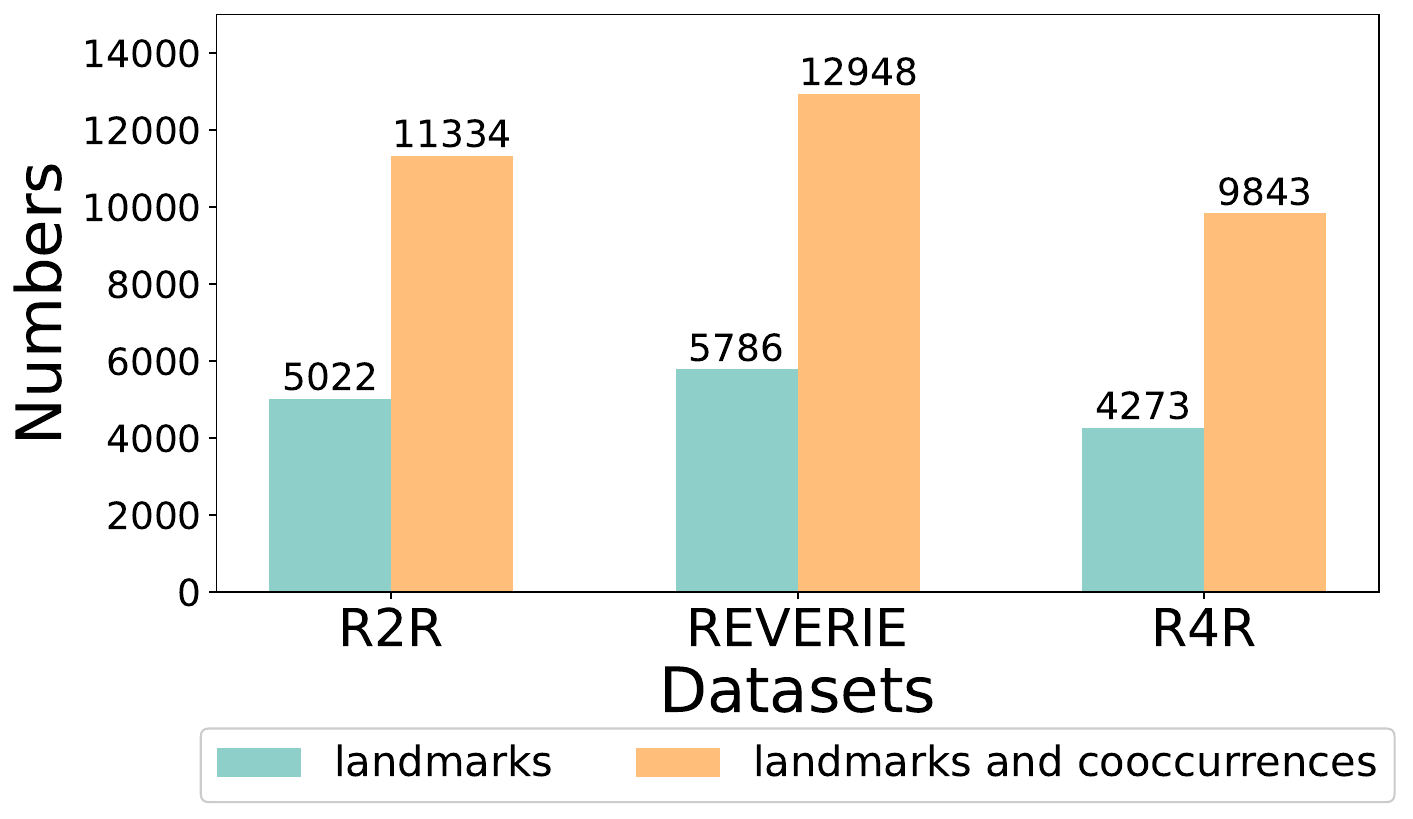}
   \caption{Numbers of Landmarks and Cooccurrences.}
   \label{fig:number}
   \vspace{-0.4cm}
\end{figure}

\noindent\textbf{Landmark Cooccurrence Generation.}
Based on the extracted navigational sequential landmarks, we can further obtain landmark cooccurrence priors from ChatGPT. Specifically, we pre-define 2 task examples for ChatGPT to generate cooccurrences. One task example followed by the prompt is as follows: 

\texttt{Tell me 3 co-occurrences of bedroom:}

\texttt{1.bed; }

\texttt{2.mirror; }

\texttt{3.nightstand; }


\texttt{Tell me 3 co-occurrences of  $U_{k}^{la}$:}

\noindent where $U_{k}^{la}$ is the $k$-th landmark in an instruction. 
We constrain ChatGPT to generate visible objects or scenes rather than abstract objects (e.g., wind) as cooccurrences by stating it explicitly in the prompt.
We only show 3 cooccurrences due to the limited space. The whole prompts with all task examples for landmark extraction and landmark cooccurrence  generation are given in the Supplementary Material.

Fig.~\ref{fig:number} presents a visualization of number comparison between landmarks and cooccurrences. From Fig.~\ref{fig:number} we can find that by introducing cooccurrences for landmark discovery, the agent can learn to align more diverse objects with observations and therefore capture rich alignment knowledge. 

\subsection{Correctable Landmark Discovery}
\label{Correctable Landmark Discovery}
Due to the lack of visual constraints, the obtained priors may be inconsistent with actual observations and cause misleading.
To address this issue, we introduce a CLIP-driven correctable landmark discovery scheme to correct the priors  in a learnable way according to actual observations, which contains the landmark shifting, the landmark discovery, and the learnable cooccurrence scoring. 
For the instruction $I$, let the landmark cooccurrence priors be $U=\{U^{la}, U^{co}\}$, where $U^{la}=\{U^{la}_{k}\}_{k=1}^{N^{la}}$ and $U^{co}=\{\{U_{1,i}^{co}\}_{i=1}^{N^{co}},..., \{U_{N^{la},i}^{co}\}_{i=1}^{N^{co}}\}$ represent the extracted landmark list and  cooccurrences, respectively. $N^{la}$ and $N^{co}$ are the numbers of landmarks and  cooccurrences, respectively.  

\subsubsection{Landmark Shifting}
At each timestep $t$, before the landmark discovery, we introduce a landmark shifting module based on CLIP to decide the current important landmark $U^{la}_{t}$ from the landmark list $U^{la}$.
Since $U^{la}$ already implies the intended order to be searched,
we keep a shifting pointer $z$ for extracting two adjacent landmarks $\{U_{z}^{la}, U_{z+1}^{la}\}$ from $U^{la}$. 
The pointer $z=1$ when $t=0$. 
The landmark shifting module predicts the current important landmark 
$U^{la}_{t}$ from $\{U_{z}^{la}, U_{z+1}^{la}\}$ based on the observations $O_{t}=\{O_{t,n}\}_{n=1}^{N_{o}}$.
For each image $B_{t,n}$ in the single-view observation $O_{t,n}$, we first use the CLIP image encoder $E^{V}_{\mathrm{CLIP}}(\cdot)$ to obtain the feature $\mathbf{f}_{B_{t,n}}$. And we obtain the current observation feature $\mathbf{f}_{B_{t}}$  by an average operation:
\vspace{-0.2cm}
\begin{equation}
\label{image feature}
\mathbf{f}_{B_{t}}=\frac{1}{N_{o}}\sum_{n}\mathbf{f}_{B_{t,n}}=\frac{1}{N_{o}}\sum_{n}{E^{V}_{\mathrm{CLIP}}(B_{t,n})}.
\end{equation}
Then, we obtain the textual features $\{\mathbf{f}_{U_{z}^{la}}, \mathbf{f}_{U_{z+1}^{la}}\}$ via the CLIP text encoder $E^{T}_{\mathrm{CLIP}}(\cdot)$:
\begin{equation}
\label{text feature}
\mathbf{f}_{U_{z}^{la}} =E^{T}_{\mathrm{CLIP}}(U_{z}^{la}),\quad\mathbf{f}_{U_{z+1}^{la}} =E^{T}_{\mathrm{CLIP}}(U_{z+1}^{la}).
\end{equation}
With $\mathbf{f}_{B_{t}}$ and $\mathbf{f}_{U_{z}^{la}}$, we obtain the probability $p_{U_{z}^{la}}$ that the current important landmark $U^{la}_{t}$ is $U_{z}^{la}$ by:
\begin{equation}
\label{current important landmark}
p_{U_{z}^{la}} = \frac{\mathrm{exp}(\mathrm{sim}(\mathbf{f}_{B_{t}},\mathbf{f}_{U_{z}^{la}})/\tau)}{\mathrm{exp}(\mathrm{sim}(\mathbf{f}_{B_{t}},\mathbf{f}_{U_{z}^{la}})/\tau)+\mathrm{exp}(\mathrm{sim}(\mathbf{f}_{B_{t}},\mathbf{f}_{U_{z+1}^{la}})/\tau)},
\end{equation}
where $\tau$ is the temperature parameter, $\mathrm{exp}(\cdot)$ represents the exponent operation, and $\mathrm{sim}(\cdot,\cdot)$ denotes the similarity computation, which is the dot product between two features. The probability $p_{U_{z+1}^{la}}$ is calculated similarly. 
Then $U^{la}_{t}=U_{z}^{la}$ if $p_{U_{z}^{la}} > p_{U_{z+1}^{la}}$ and $U^{la}_{t}=U_{z+1}^{la}$ if $p_{U_{z}^{la}} \le p_{U_{z+1}^{la}}$. If $U^{la}_{t}=U_{z+1}^{la}$, the pointer $z$ will move to $z+1$ at timestep $t+1$. Due to the scene complexity, the pointer $z$ may not move during many navigation steps. Therefore, we force the movement of $z$ when it continuously points to the same position (see Fig.~\ref{fig:visualiation} for example). Specifically, we define a counter to record the number of no-shifting steps. If the number of no-shifting steps exceeds a pre-defined step threshold, we force the pointer $z$ to move to $z+1$ and reset the counter to 0. The step threshold is set to 1 when an instruction contains more than 3 landmarks. Otherwise, the step threshold is set as 2.

Note that Hong et al. also introduce a sub-instruction shifting module~\cite{Hong2020SubInstructionAV} which is learnable to facilitate the alignment between instructions and trajectories. Different from~\cite{Hong2020SubInstructionAV}, we do not add learnable parameters in our heuristic landmark shifting module and instead construct a learnable cooccurrence scoring module (Sec.~\ref{Learnable Cooccurrence Scoring}) that effectively facilitates cooccurrence-based correctable landmark discovery. Moreover, we conduct fine-grained shifting among landmarks rather than sub-instructions to enable sequential landmark discovery.


\subsubsection{Landmark Discovery}
In discrete VLN task, the actions are chosen from candidate single-view observations. Therefore, to adapt our landmark discovery framework for navigation action prediction, we transform the landmark discovery as locating the landmark in one of the candidate observations, which is consistent with the action prediction process. With the current important landmark $U^{la}_{t}$ and its cooccurrences $\{U^{co}_{t,i}\}_{i=1}^{N^{co}}$, we conduct the landmark discovery through CLIP to highlight the single-view observation where  $U^{la}_{t}$ with $\{U^{co}_{t,i}\}_{i=1}^{N^{co}}$ is most likely to appear for assisting action decision.
To this end, for each  single-view observation $O_{t,n}$, we calculate the probability $p_{U^{la}_{t,n}}$ that the current important landmark $U^{la}_{t}$ appears in it by:
\vspace{-0.2cm}
\begin{equation}
\label{landmark discovery}
p_{U^{la}_{t,n}} = \frac{\mathrm{exp}(\mathrm{sim}(\mathbf{f}_{U^{la}_{t}}, \mathbf{f}_{B_{t,n}})/\tau)}{\sum_{m=1}^{N_{o}}\mathrm{exp}(\mathrm{sim}(\mathbf{f}_{U^{la}_{t}}, \mathbf{f}_{B_{t,m}})/\tau)},
\end{equation}
where $\mathbf{f}_{U_{t}^{la}}=\mathbf{f}_{U_{z}^{la}}$ or $\mathbf{f}_{U_{z+1}^{la}}$,  $\mathbf{f}_{B_{t,n}}$ is the visual feature  for $O_{t,n}$ (Eq.~\ref{image feature}), $\mathrm{exp}(\cdot)$ represents the exponent operation, and $\mathrm{sim}(\cdot,\cdot)$ denotes the similarity computation, which is the dot product between two features. 
We also calculate the probability $p_{U^{co}_{t,i,n}}$ that $U^{co}_{t,i}$ appears in  $O_{t,n}$ for each cooccurrence $U^{co}_{t,i}$ like Eq.~\ref{landmark discovery}. 

\subsubsection{Learnable Cooccurrence Scoring}
\label{Learnable Cooccurrence Scoring}
Since the cooccurrence priors are extracted without visual constraints, the  cooccurrences may not appear in the observations or may appear in multiple single-view observations to cause misleading. 
Therefore, we introduce a learnable cooccurrence scoring module to correct the priors according to actual observations.
Moreover, due to the complexity of the scene and the instruction, it is hard for the large model to determine the correct landmark totally correctly. Through the introduction of the learnable cooccurrence scoring module trained together with the baseline agent, the noise brought by the landmark shifting module and the landmark discovery module can be effectively mitigated.

Specifically, we obtain the landmark score $s^{la}$ and the cooccurrence score $\{s_{i}^{co}\}_{i=1}^{N^{co}}$ through a dynamic updated state feature $\mathbf{f}_{S_{t}}$. Then we use $s^{la}$ and $s_{i}^{co}$ to obtain the corrected landmark prediction probability $\tilde{p}_{U_{t,n}}$ based on the  probabilities $p_{U^{la}_{t,n}}$ and $\{\{p_{U^{co}_{t,1,n}}\}_{n=1}^{N_{o}},...,\{p_{U^{co}_{t,N^{co},n}}\}_{n=1}^{N_{o}}\}$.
Through $s^{la}$ and $s_{i}^{co}$, the important cooccurrence in the priors can be highlighted while the unimportant ones are suppressed by $\tilde{p}_{U_{t,n}}$.
At timestep $t$, we calculate 
$\mathbf{f}_{S_{t}}$ based on the instruction feature $\mathbf{f}_{I}$ (Eq.~\ref{unimodal features}) and the current observation feature $\mathbf{f}_{B_{t}}$ (Eq.~\ref{image feature}):
\begin{equation}
\mathbf{f}_{S_{t}}= E^{l}([\mathbf{f}_{I}^{cls};\mathbf{f}_{B_{t}}]), 
\end{equation}
where $\mathbf{f}_{I}^{cls}$ is the instruction feature of the [CLS] token in the baseline~\cite{Chen2021HistoryAM,Chen2022ThinkGA}, and $E^{l}(\cdot)$  contains a linear layer, a ReLU activation function, a layer normalization function,  and a Dropout operation.  Then, we obtain the cooccurrence score $s_{i}^{co}$ for cooccurrence $U^{co}_{t,i}$ ($1\le i\le N^{co}$) by:
\begin{equation}
\label{cooccurrence score}
s_{i}^{co} = \mathrm{sim}(\mathbf{f}_{S_{t}}, \mathbf{f}_{U^{co}_{t,i}}),
\end{equation}
where $\mathrm{sim}(\cdot,\cdot)$ denotes the similarity computation which is the dot product between two features and $\mathbf{f}_{U^{co}_{t,i}}=E^{T}_{\mathrm{CLIP}}(U^{co}_{t,i})$ is the textual feature. We also calculate the score $s^{la}$ for the current important landmark $U^{la}_{t}$ as Eq.~\ref{cooccurrence score}. Based on the original  probability $p_{U^{la}_{t,n}}$ and $\{p_{U^{co}_{t,i,n}}\}_{i=1}^{N^{co}}$ 
, we obtain the corrected landmark prediction probability $\tilde{p}_{U_{t,n}}$ for the single-view observation $O_{t,n}$ by:
\vspace{-0.2cm}
\begin{equation}
\label{correction}
\tilde{p}_{U_{t,n}}=p_{U^{la}_{t,n}}\cdot s^{la}+\sum_{i=1}^{N^{co}}p_{U^{co}_{t,i,n}}\cdot s_{i}^{co},
\end{equation}

\noindent\textbf{Consistency Loss.} 
To constrain the optimization of the learnable cooccurrence scoring module,
we introduce a consistency loss $\mathcal{L}_{\mathrm{cs}}$, which is the cross-entropy loss and  calculated based on the corrected landmark prediction probability $\tilde{p}_{U_{t,n}}$ and the ground-truth action $g_{t}$:
\begin{equation}
\mathcal{L}_{\mathrm{cs}}=-\sum_{t}g_{t}\mathrm{log}(\mathbf{\tilde{p}}_{U_{t}}),
\end{equation}
where $\mathbf{\tilde{p}}_{U_{t}}=\{\tilde{p}_{U_{t,n}}\}_{n=1}^{N_{o}}$. 
Since both action prediction and landmark discovery aim at selecting the ground-truth candidate observation, by enforcing the consistency between  $\mathbf{\tilde{p}}_{U_{t}}$ and $g_{t}$,  
the scoring module can learn to correct the importance of cooccurrences for accurate landmark discovery.
As shown in Fig.~\ref{fig:overview}, through the learnable cooccurrence scoring module under the constraint of the consistency loss $\mathcal{L}_{\mathrm{cs}}$, the cooccurrence ``counter'' is significantly highlighted to indicate the ground-truth action.


\subsection{Observation Enhancement}
\label{Observation Enhancing}
To encourage VLN agents to learn sufficient open-world alignment knowledge under the correctable landmark discovery framework, we design an observation enhancement strategy, 
where we use the corrected landmark feature  $\mathbf{f}_{U_{t}}=\{\mathbf{f}_{U_{t,n}}\}_{n=1}^{N_{o}}$ and the original observation feature $\mathbf{f}_{O_{t}}=\{\mathbf{f}_{O_{t,n}}\}_{n=1}^{N_{o}}$ to obtain the enhanced observation features $\mathbf{f'}_{O_{t}}=\{\mathbf{f'}_{O_{t,n}}\}_{n=1}^{N_{o}}$ for  action decision.
Specifically, at timestep $t$, we calculate $\mathbf{f}_{U_{t,n}}$ by:
\vspace{-0.2cm}
\begin{equation}
\mathbf{f}_{U_{t,n}} =p_{U^{la}_{t,n}}\cdot s^{la}\cdot\mathbf{f}_{U^{la}_{t}}+\sum_{i=1}^{N^{co}}p_{U^{co}_{t,i,n}}\cdot s_{i}^{co}\cdot\mathbf{f}_{U^{co}_{t,i}},
\end{equation}
where $\mathbf{f}_{U^{la}_{t}}$ and $\mathbf{f}_{U^{co}_{t,i}}$ are the textual features of the landmark $U^{la}_{t}$ and the cooccurrence $U^{co}_{t,i}$. 
Through $s^{la}$ and $s_{i}^{co}$, the feature of the important cooccurrence can be highlighted.
Then, we use $\mathbf{f}_{U_{t,n}}$ and  $\mathbf{f}_{O_{t,n}}$ to obtain the enhanced observation feature $\mathbf{f'}_{O_{t,n}}$ for $O_{t,n}$:
\begin{equation}
\mathbf{f'}_{O_{t,n}}=\mathbf{f}_{O_{t,n}}+\mathbf{f}_{U_{t,n}}.
\end{equation}
The enhanced observation features $\mathbf{f'}_{O_{t}}=\{\mathbf{f'}_{O_{t,n}}\}_{n=1}^{N_{o}}$ are finally utilized for action decision. 

\noindent\textbf{Contrastive Loss.}
For improving the modality alignment and encouraging the enhanced observation feature $\mathbf{f'}_{O_{t,n}}$ to be more discriminative for facilitating action decision, we introduce a contrastive loss $\mathcal{L}_{\mathrm{ct}}$. Through $\mathcal{L}_{\mathrm{ct}}$, we enforce the observation feature $\mathbf{f}_{O_{t,n}}$ to be close with the paired corrected landmark feature $\mathbf{f}_{U_{t,n}}$ while far away from the non-paired corrected landmark feature $\overline{\mathbf{f}}_{U_{t,n}}=\{\mathbf{f}_{U_{t,j}}\}$ ($1\le j\le N_{o}$ and $j\neq n$). $\mathcal{L}_{\mathrm{ct}}$ is composed of two losses $\mathcal{L}_{\mathrm{ct}}^{o\rightarrow{u}}$ and $\mathcal{L}_{\mathrm{ct}}^{u\rightarrow{o}}$. We calculate $\mathcal{L}_{\mathrm{ct}}^{o\rightarrow{u}}$ by: 
\begin{equation}
\mathcal{L}_{\mathrm{ct}}^{o\rightarrow{u}}=-\sum_{t,n}\mathrm{log}(\frac{\mathrm{e}^{\mathrm{sim}(\mathbf{f}_{O_{t,n}}, \mathbf{f}_{U_{t,n}})/\tau}}{\mathrm{e}^{\mathrm{sim}(\mathbf{f}_{O_{t,n}}, \mathbf{f}_{U_{t,n}})/\tau}+\sum\limits_{\overline{\mathbf{f}}_{U_{t,n}}}\mathrm{e}^{\mathrm{sim}(\mathbf{f}_{O_{t,n}}, \overline{\mathbf{f}}_{U_{t,n}})/\tau}}),
\end{equation}
where $\mathrm{sim}(\cdot,\cdot)$ denotes the similarity computation, which is the dot product between two features. $\mathcal{L}_{\mathrm{ct}}^{u\rightarrow{o}}$ is calculated in a symmetric way as $\mathcal{L}_{\mathrm{ct}}^{o\rightarrow{u}}$. Then we obtain the  contrastive loss  $\mathcal{L}_{\mathrm{ct}}$ by $\mathcal{L}_{\mathrm{ct}}=\frac{1}{2}(\mathcal{L}_{\mathrm{ct}}^{o\rightarrow{u}}+\mathcal{L}_{\mathrm{ct}}^{u\rightarrow{o}})$.

\subsection{Action Prediction}
\label{Action Prediction}
The action $\mathbf{\tilde{a}}_{t}$ is predicted based on the enhanced observation feature $\mathbf{f'}_{O_{t}}$ and the instruction feature $\mathbf{f}_{I}$ through the cross-modal Transformer encoder $E^{c}(\cdot)$ and the action prediction module $E^{a}(\cdot)$ (see Eq.~\ref{eq:ec} and Eq.~\ref{eq:ea}).
The total training objective $\mathcal{L}$ of CONSOLE is:
\begin{equation}
\mathcal{L}=\mathcal{L}_{\mathrm{nav}}+\lambda_{1}\mathcal{L}_{\mathrm{cs}}+\lambda_{2}\mathcal{L}_{\mathrm{ct}},
\end{equation}
where $\lambda_{1}$ and $\lambda_{2}$ are the balance meta-parameters.

\section{Experiments}

\begin{table*}[t]

\caption{Comparison with SOTA methods on R2R. * denotes using additional large-scale unannotated data. \textbf{Bold} and \textcolor{blue}{ \textbf{Blue}} denote the best and runner-up results.}
	\vspace{-0.2cm}
	\label{tab:com with sota}
	\resizebox{1.0\linewidth}{!}{
	{\renewcommand{\arraystretch}{1.0}
		\begin{tabular}{c||c|c|c|c|c|c|c|c|c|c|c|c}

			\specialrule{.1em}{.05em}{.05em}
			\multirow{2}{*}{Method}&\multicolumn{4}{c|}{Val Seen }&\multicolumn{4}{c|}{Val Unseen}&\multicolumn{4}{c}{Test Unseen}\cr\cline{2-13}
			&TL&NE $\downarrow$&SR $\uparrow$&SPL $\uparrow$&TL&NE $\downarrow$&SR $\uparrow$&SPL $\uparrow$&TL&NE $\downarrow$&SR $\uparrow$&SPL $\uparrow$\cr
			\hline
			
        
            Seq2Seq~\cite{anderson2018vision}&11.33&6.01&39&-&8.39&7.81&22&-&8.13&7.85&20&18\\
            RCM+SIL(train) \cite{wang2019reinforced}&10.65&3.53&67&-&11.46&6.09&43&-&11.97&6.12&43&38\\
            
   
          EnvDropout~\cite{tan2019learning}&11.00&3.99&62&59&10.70&5.22&52&48&11.66&5.23&51&47\\  
            RelGraph \cite{hong2020language}&10.13&3.47&67&65&9.99&4.73&57&53&10.29&4.75&55&52\\

        PREVALENT~\cite{hao2020towards} & 10.32 & 3.67 & 69 & 65 & 10.19 & 4.71 & 58 & 53 & 10.51 & 5.30 & 54 & 51 \\
		 ORIST~\cite{Qi2021TheRT}&-&-&-&-&10.90&4.72&57&51&11.31&5.10&57&52\\ VLN$\circlearrowright$BERT~\cite{hong2021vln}&11.13&2.90&72&68&12.01&3.93&63&57&12.35&4.09&63&57\\
 HOP~\cite{Qiao2022HOPHA}&11.51&2.46&76&70&12.52&3.79&64&57&13.29&3.87&64&58\\
 VLN-SIG~\cite{Li2023ImprovingVN}&-&-&-&-&-&-&72&62&-&-&\textcolor{blue}{\textbf{72}}&60\\
 KERM~\cite{Li2023KERMKE}&12.16&2.19&80&74&13.54&3.22&72&61&14.60&3.61&70&59\\
 Meta-Explore~\cite{Hwang2023MetaExploreEH}&11.95&\textbf{2.11}&\textbf{81}&\textbf{75}&13.09&3.22&72&62&14.25&3.57&71&\textcolor{blue}{\textbf{61}}\\
 ScaleVLN*~\cite{Wang2023ScalingDG}&13.24&2.12&\textbf{81}&\textbf{75}&14.09&\textbf{2.09}&\textbf{81}&\textbf{70}&13.92&\textbf{2.27}&\textbf{80}&\textbf{70}\\
 NavGPT~\cite{Zhou2023NavGPTER}&-&-&-&-&11.45&6.46&34&29&-&-&-&-\\
		 \hline HAMT~\cite{Chen2021HistoryAM} (baseline)&11.15&2.51&76&72&11.46&3.62&66&61&12.27&3.93&65&60\\ 
   HAMT+CONSOLE (ours)&11.73&2.65&74&70&11.87&3.39&69&\textcolor{blue}{\textbf{63}}&12.42&3.78&66&\textcolor{blue}{\textbf{61}}\\
    DUET~\cite{Chen2022ThinkGA} (baseline)&12.32&2.28&79&73&13.94&3.31&72&60&14.73&3.65&69&59\\
    DUET+CONSOLE (ours)&12.74&2.17&79&73&13.59&\textcolor{blue}{\textbf{3.00}}&\textcolor{blue}{\textbf{73}}&\textcolor{blue}{\textbf{63}}&14.31&\textcolor{blue}{\textbf{3.30}}&\textcolor{blue}{\textbf{72}}&\textcolor{blue}{\textbf{61}}\\
 \specialrule{.1em}{.05em}{.05em}

		\end{tabular}}}
\end{table*}

\subsection{Experimental Setup}
\textbf{Datasets.}
We evaluate CONSOLE on four mainstream VLN benchmarks: R2R~\cite{anderson2018vision}, REVERIE~\cite{qi2020reverie}, R4R~\cite{jain2019stay}, and RxR~\cite{ku2020room}. R2R contains 90 indoor scenes with 7189 trajectories. 
REVERIE replaces the fine-grained instructions in R2R with high-level instructions to locate remote objects. R4R concatenates two adjacent tail-to-head trajectories in R2R, forming longer instructions and trajectories. RxR contains much more complex instructions and trajectories than R2R. Since CLIP~\cite{radford2021learning} is pretrained on English language data, we use the English subset of RxR (both en-IN and en-US) for verification, which includes 26464, 2939, 4551 path-instruction pairs for Training, Val Seen, and Val Unseen, respectively.

\noindent\textbf{Evaluation Metrics.}
The evaluation metrics for R2R~\cite{anderson2018vision} are: 1) Trajectory Length (TL): the average length of the agent's navigated path in meters, 2) Navigation Error (NE): the average distance in meters between the agent's destination and the target viewpoint, 3) Success Rate (SR): the ratio of success, where the agent stops within three meters of the target point, and 4) Success rate weighted by Path Length (SPL): success rate normalized by the ratio between the length of the shortest path and the predicted path. For R4R~\cite{jain2019stay} and RxR~\cite{ku2020room}, we further employ 5) the Coverage weighted by Length Score (CLS), 6) the normalized Dynamic Time Warping (nDTW), and 7) the Success weighted by nDTW (SDTW) for measuring the path fidelity. For REVERIE~\cite{qi2020reverie}, three other metrics are used: 8) Remote Grounding Success Rate (RGS): the ratio of grounding the correct object when stopping, 9) Remote Grounding Success rate weighted by Path Length (RGSPL): weight RGS by TL, and 10) Oracle Success Rate (OSR): the ratio of containing a viewpoint along the path where the target object is visible.

\noindent\textbf{Implementation Details.} We conduct
all experiments on an NVIDIA 3090 GPU. The batch size is set to 4 and the model is trained for 300K iterations on all datasets. The navigation loss of CONSOLE is the same as that in each baseline. The temperature parameter $\tau$ is set as 0.5 empirically. We use the SGD optimizer with the learning rate 0.1 for training $E^{l}(\cdot)$ in the learnable cooccurrence scoring module. The loss weights $\lambda_{1}$ and $\lambda_{2}$ are set as 0.1 on all datasets. The same augmented data in \cite{Chen2021HistoryAM} is used for R2R for fair comparison.
To sufficiently utilize the navigation ability of existing VLN agents, we train CONSOLE using the pretrained checkpoint of the baseline.

For facilitating implementation, we use the CLIP visual features (ViT-B/32) released by~\cite{Chen2021HistoryAM} instead of using the CLIP image encoder. 
And we pre-extract the landmark cooccurrence priors before training. 
To better activate the landmark discovery capability of CLIP, 
we add the prompt ``a photo of a'' like~\cite{radford2021learning} to obtain the CLIP textual feature for a specific landmark during the landmark discovery. 
For the landmark extraction, we use
the same task examples for R2R, R4R and RxR since the instructions for them are all fine-grained. For REVERIE which
have high-level instructions, we pre-define  different task
examples.
Since the number of landmarks in R2R augmented data is $\sim$75 times larger than that of R2R Train split, using ChatGPT directly for cooccurrence extraction has low feasibility  regarding both time and cost. Alternatively, we obtain the cooccurrences for the augmented data through a pre-built landmark-cooccurrence vocabulary based on the landmark cooccurrence priors of R2R Train split. We compute the most semantically similar landmark in the vocabulary for each landmark in R2R augmentation data to ensure the quality of the extracted cooccurrences.
For each path in R4R, we directly combine the landmarks and cooccurrences from 2 corresponding R2R instructions for improving efficiency.


\begin{table*}[t]
	\fontsize{20}{20}\selectfont

\caption{Navigation and object grounding performance on REVERIE. \textbf{Bold} and \underline{underline} denote the best and runner-up results.}
	\vspace{-0.2cm}
	\label{tab:com with sota on reverie}
	\resizebox{1.0\linewidth}{!}{
	{\renewcommand{\arraystretch}{1.2}
		\begin{tabular}{c||c|c|c|c|c|c|c|c|c|c|c|c|c|c|c}

			\specialrule{.1em}{.05em}{.05em}
			\multirow{2}{*}{Method}&\multicolumn{5}{c|}{Val Seen}&\multicolumn{5}{c|}{Val Unseen}&\multicolumn{5}{c}{Test Unseen}\cr\cline{2-16}
			&SR $\uparrow$&OSR$\uparrow$&SPL $\uparrow$&RGS$\uparrow$&RGSPL$\uparrow$&SR $\uparrow$&OSR$\uparrow$&SPL $\uparrow$&RGS$\uparrow$&RGSPL$\uparrow$&SR $\uparrow$&OSR$\uparrow$&SPL $\uparrow$&RGS$\uparrow$&RGSPL$\uparrow$\cr
			\hline
			
        
            Seq2Seq~\cite{anderson2018vision}&29.59&35.70&24.01&18.97&14.96&4.20&8.07&2.84&2.16&1.63&3.99&6.88&3.09&2.00&1.58\\
            RCM~\cite{wang2019reinforced}&23.33&29.44&21.82&16.23&15.36&9.29&14.23&6.97&4.89&3.89&7.84&11.68&6.67&3.67&3.14\\
           
          SMNA~\cite{ma2019self}&41.25&43.29&39.61&30.07&28.98&8.15&11.28&6.44&4.54&3.61&5.80&8.39&4.53&3.10&2.39\\
         FAST-MATTN~\cite{qi2020reverie}&50.53&55.17&45.50&31.97&29.66&14.40&28.20&7.19&7.84&4.67&19.88&30.63&11.60&11.28&6.08\\ 
    SIA~\cite{Lin2021SceneIntuitiveAF}&61.91&65.85&57.08&45.96&42.65&31.53&44.67&16.28&22.41&11.56&30.80&44.56&14.85&19.02&9.20\\
            VLN$\circlearrowright$BERT~\cite{hong2021vln}&51.79&53.90&47.96&38.23&35.61&30.67&35.02&24.90&18.77&15.27&29.61&32.91&23.99&16.50&13.51\\
    HOP~\cite{Qiao2022HOPHA}&54.81&56.08&48.05&40.55&35.79&30.39&35.30&25.10&18.23&15.31&29.12&32.26&23.37&17.13&13.90\\
    TD-STP~\cite{Zhao2022TargetDrivenST}&-&-&-&-&-&34.88&39.48&27.32&21.16&16.56&35.89&40.26&27.51&19.88&15.40\\
    KERM~\cite{Li2023KERMKE}&71.89&74.49&64.04&57.55&51.22&49.02&53.65&\textbf{34.83}&\underline{33.97}&\textbf{24.14}&52.26&57.44&\textbf{37.46}&\underline{32.69}&\textbf{23.15}\\
    

    \hline
    DUET~\cite{Chen2022ThinkGA} &71.75&73.86&63.94&57.41&51.14&46.98&51.07&33.73&32.15&23.03&52.51&56.91&36.06&31.88&22.06\\
    CONSOLE (ours)&\textbf{74.14}&\textbf{76.25}&\textbf{65.15}&\textbf{60.08}&\textbf{52.69}&\textbf{50.07}&\textbf{54.25}&\underline{34.40}&\textbf{34.05}&\underline{23.33}&\textbf{55.13}&\textbf{59.60}&\underline{37.13}&\textbf{33.18}&\underline{22.25}\\
 \specialrule{.1em}{.05em}{.05em}

			\end{tabular}}}
\end{table*}

\subsection{Quantitative Results}

\begin{table}
    \caption{Comparison on R4R val unseen split.\\ * denotes the results of our re-implementation.}
    
    \small
    \centering
    \resizebox{\linewidth}{!}{
    \begin{tabular}{l|ccccc}
    \specialrule{.1em}{.05em}{.05em}
	Methods&NE$\downarrow$&SR$\uparrow$&CLS$\uparrow$&nDTW$\uparrow$&SDTW$\uparrow$\\		
   
			\hline
			SF~\cite{fried2018speaker}&8.47&24&30&-&-\\
   RCM~\cite{wang2019reinforced}&-&29&35&30&13\\
   PTA~\cite{landi2019perceive}&8.25&24&37&32&10\\
   EGP~\cite{deng2020evolving}&8.0&30.2&44.4&37.4&17.5\\
   RelGraph~\cite{hong2020language}&7.43&36&41&47&34\\
   
        VLN$\circlearrowright$BERT~\cite{hong2021vln}&6.67&43.6&51.4&45.1&29.9\\
        
        HAMT~\cite{Chen2021HistoryAM} &6.09&\textbf{44.6}&57.7&50.3&31.8\\
        \hline
        HAMT*~\cite{Chen2021HistoryAM}&6.25&41.50&58.56&51.50&30.69\\ 
        CONSOLE (ours)&\textbf{5.99}&43.22&\textbf{61.45}&\textbf{54.34}&\textbf{32.72}\\
 \specialrule{.1em}{.05em}{.05em}
    \end{tabular}
}
\label{tab:com on r4r}
\vspace{-0.2cm}
\end{table}

\begin{table}
    \caption{Comparison on RxR-En val unseen split.\\ * denotes the results of our re-implementation.}
    
    \small
    \centering
    \resizebox{\linewidth}{!}{
    \begin{tabular}{l|ccccc}
    \specialrule{.1em}{.05em}{.05em}
	Methods&SR$\uparrow$&SPL$\uparrow$&CLS$\uparrow$&nDTW$\uparrow$&SDTW$\uparrow$\\
 \hline
   EnvDrop~\cite{tan2019learning}&38.5&34&54&51&32\\
   Syntax~\cite{li2021improving}&39.2&35&56&52&32\\
   HAMT*~\cite{Chen2021HistoryAM}&43.20&40.41&59.57&55.52&36.39\\
   CONSOLE (ours)&\textbf{46.34}&\textbf{42.92}&\textbf{60.85}&\textbf{57.11}&\textbf{38.85}\\
   
 \specialrule{.1em}{.05em}{.05em}
    \end{tabular}
}
\label{tab:com on rxr}
\vspace{-0.2cm}
\end{table}

\noindent\textbf{Comparison with SOTAs.} 
Table~\ref{tab:com with sota}\footnote{The original value 2.29 of NE under Val Unseen in HAMT is a typo, which is actually 3.62 and confirmed by one of HAMT's authors.}, Table~\ref{tab:com with sota on reverie}, Table~\ref{tab:com on r4r}, and Table~\ref{tab:com on rxr} show the performance comparison between recent approaches and CONSOLE, where we can find that CONSOLE outperforms the strong baselines HAMT~\cite{Chen2021HistoryAM} and DUET~\cite{Chen2022ThinkGA} consistently on R2R, REVERIE, R4R, and RxR especially in unseen scenarios (e.g., CONSOLE outperforms DUET in SPL by 3\% and 2\% under Val Unseen and Test Unseen on R2R, respectively). 
The results demonstrate the strong generalization ability of CONSOLE for diverse instructions and scenarios. 
Moreover, CONSOLE establishes new SOTA results on R2R and R4R under unseen scenes.
CONSOLE also shows good performance in seen scenarios on different datasets, e.g., it significantly outperforms DUET in Val Seen on REVERIE.
All these results show the effectiveness of CONSOLE, demonstrating that CONSOLE successfully activates and utilizes open-world knowledge from large models for assisting VLN. 

Note that although the concurrent work ScaleVLN~\cite{Wang2023ScalingDG} outperforms previous works as well as CONSOLE, it leverages significantly larger data resources and therefore a direct comparison would not be fair. Another concurrent work, NavGPT~\cite{Zhou2023NavGPTER}, introduces LLM to serve as the navigation decision backbone directly. Different from NavGPT~\cite{Zhou2023NavGPTER}, our CONSOLE combines the advantages of both LLM and existing vision-based VLN models to improve the navigation performance.


\begin{table*}[t]
	\fontsize{7}{7}\selectfont

\caption{Ablation study of the learnable cooccurrence scoring module (LS) and the observation enhancement module (OE) for different baselines (HAMT~\cite{Chen2021HistoryAM} \& DUET~\cite{Chen2022ThinkGA}) and datasets (R2R~\cite{anderson2018vision} \& REVERIE~\cite{qi2020reverie}). }
	\label{tab:LS and OE}
	\resizebox{1.0\linewidth}{!}{
	{\renewcommand{\arraystretch}{1.1}
		\begin{tabular}{c||c|c|c|c|c|c|c|c|c|c|c}

			\specialrule{.1em}{.05em}{.05em}
			\multirow{2}{*}{Method}&\multicolumn{3}{c|}{R2R-HAMT }&\multicolumn{3}{c|}{R2R-DUET}&\multicolumn{5}{c}{REVERIE-DUET}\cr\cline{2-12}
			&NE $\downarrow$&SR $\uparrow$&SPL $\uparrow$&NE $\downarrow$&SR $\uparrow$&SPL $\uparrow$&SR $\uparrow$&OSR$\uparrow$&SPL $\uparrow$&RGS$\uparrow$&RGSPL$\uparrow$\cr
			\hline
			
        
      Baseline&3.62&66&61&3.31&72&60&46.98&33.73&51.07&32.15&23.03\\
      \hline
      No LS \& No OE&3.53&67.39&62.42&3.12&72.54&60.64&48.68&33.27&\textbf{54.96}&32.83&22.53\\
      LS \& No OE&3.46&68.03&62.59&3.07&73.35&61.51&49.42&34.33&53.82&33.51&23.32\\
      LS \& OE&\textbf{3.39}&\textbf{68.62}&\textbf{63.06}&\textbf{3.00}&\textbf{73.44}&\textbf{63.02}&\textbf{50.07}&\textbf{34.40}&54.25&\textbf{34.05}&\textbf{23.33}\\
 \specialrule{.1em}{.05em}{.05em}

		\end{tabular}}}
	\vspace{-0.2cm}
\end{table*}

\begin{table}
    \caption{Ablation Studies of cooccurrence number, landmark shifting module, feature fusion strategies, loss weights, and temperature parameter on R2R.}
\vspace{-0.2cm}
\fontsize{4}{4}\selectfont
    \centering
    \resizebox{\linewidth}{!}{
    {\renewcommand{\arraystretch}{1.2}
    \begin{tabular}{l|ccc}
    \specialrule{.1em}{.05em}{.05em}
			\multirow{2}{*}{Method}&\multicolumn{3}{c}{Val Unseen}\cr\cline{2-4}
   &NE$\downarrow$&SR$\uparrow$&SPL$\uparrow$\cr

			\hline
			
        Baseline~\cite{Chen2021HistoryAM}&3.62&66&61\\
        \hline
        \multicolumn{3}{l}{$\triangleright$  Cooccurrence number:}\cr\cline{1-4}
        \hline
        num=0&3.60&66.84&61.78\\
        num=1&3.60&66.45&61.62\\
        num=5&3.53&67.39&62.42\\
        num=10&3.52&67.31&62.19\\
        \hline
        \multicolumn{3}{l}{$\triangleright$  Landmark shifting:}\cr\cline{1-4}
        \hline
        Global&3.59&66.88&61.51\\
        Noforcedshift&3.54&67.22&62.05\\
        Shift&3.53&67.39&62.42\\
 \hline
 \multicolumn{3}{l}{$\triangleright$ strategies for fusing landmark features and observation features:}\cr\cline{1-4}
 \hline
 concatenation&3.61&66.24&61.12\\
 summation&3.46&68.03&62.59\\
 \hline
        \multicolumn{3}{l}{$\triangleright$ Loss weights:}\cr\cline{1-4}
        \hline
        $\lambda_{1}$=0.1 \& $\lambda_{2}$=1&3.50&67.48&62.56\\
        $\lambda_{1}$=1 \& $\lambda_{2}$=0.1&3.63&67.35&62.53\\
           $\lambda_{1}$=1 \& $\lambda_{2}$=1&3.58&67.65&62.26\\
              $\lambda_{1}$=0.1 \& $\lambda_{2}$=0.1 (ours)&\textbf{3.39}
&\textbf{68.62}&\textbf{63.06}\\
\hline
\multicolumn{3}{l}{$\triangleright$ Temperature parameter $\tau$:}\cr\cline{1-4}
\hline
$\tau$=0.05&3.48&67.56&62.36\\
$\tau$=0.1&3.44&68.03&62.71\\
$\tau$=1&3.50&68.11&\textbf{63.10}\\
$\tau$=0.5 (ours)&\textbf{3.39}&\textbf{68.62}&\underline{63.06}\\

 \specialrule{.1em}{.05em}{.05em}
    \end{tabular}
}}
\label{tab:ablation}
\vspace{-0.4cm}
\end{table}

\begin{table}
    \caption{Comparison of LLMs on R2R. The baseline we choose is DUET~\cite{Chen2022ThinkGA}.}
\vspace{-0.2cm}
\fontsize{3}{3}\selectfont
    \centering
    \resizebox{\linewidth}{!}{
    {\renewcommand{\arraystretch}{1.1}
    \begin{tabular}{l|ccc}
    \specialrule{.1em}{.05em}{.05em}
			\multirow{2}{*}{Method}&\multicolumn{3}{c}{Val Unseen}\cr\cline{2-4}
   &NE$\downarrow$&SR$\uparrow$&SPL$\uparrow$\cr

			\hline
			
        Baseline~\cite{Chen2021HistoryAM}&3.31&72&60\\
        \hline
        Spacy&3.11&73.05&61.33\\
        GPT-3&\textbf{3.00}&\textbf{73.82}&61.56\\
        Vicuna7B&3.06&73.31&61.65\\
        Vicuna13B&3.07&73.39&62.14\\
              ChatGPT (ours)&\textbf{3.00}
&73.44&\textbf{63.02}\\
 \specialrule{.1em}{.05em}{.05em}
    \end{tabular}
}}
\label{tab:LLMs}
\vspace{-0.4cm}
\end{table}

\noindent\textbf{Ablation Study.} 
To further analyze the effects of different components in CONSOLE, we conduct extensive ablation studies, and the results are given in Table~\ref{tab:LS and OE} and Table~\ref{tab:ablation}. Specifically, to analyze   the effects of two core modules in CONSOLE, i.e., the learnable cooccurrence scoring module (LS) and the observation enhancement module (OE), we give the ablation study results of them for different baselines and datasets in Table~\ref{tab:LS and OE}. 
Concretely, ``LS \& No OE" outperforms ``No LS \& No OE" by generally ~0.8\% improvement on different baselines and datasets, which shows the effectiveness of learnable cooccurrence scoring module. For example, ``LS \& No OE" outperforms ``No LS \& No OE" by 1.06\% and 0.87\% on SPL for REVERIE-DUET and R2R-DUET, respectively. Similarly, 
``LS \& OE" outperforms  ``LS \& No OE" on all metrics on different baselines and datasets, which also shows the effectiveness and the generalization ability of the observation enhancement module. 
The results in Table~\ref{tab:LS and OE} sufficiently show the effectiveness of two core modules in CONSOLE.

We further conduct the ablation studies for the cooccurrence number, landmark shifting module, 
strategies for fusing the corrected landmark features and the original observation features, loss weights, and temperature parameter, and the results are presented in Table~\ref{tab:ablation}.
From Table~\ref{tab:ablation} we can draw into the following conclusions:
1) The results of ``Cooccurrence number'' show that the landmark cooccurrence priors greatly facilitate the landmark discovery, e.g., ``num=5" outperforms ``num=0" by 0.55\% and 0.64\% in SR and SPL, respectively. Moreover, the results of ``num=5,10" show that adding too many co-occurrences may not bring further performance improvement. It is reasonable since too many co-occurrences may introduce unexpected noise to mislead the agent in the meanwhile.
2) The results of ``Landmark Shifting'' show that landmark shifting module is helpful for accurate landmark discovery. Especially, ``Shift" outperforms ``Global" by 0.91\% in SPL.
3) The results of ``strategies for fusing the corrected landmark features and the original observation features'' show that ``summation'' outperforms ``concatenation'' and ``concatenation'' is comparable with the baseline. This result reveals that ``summation" is a more effective way for fusing the corrected landmark features and the original observation features.
4) The results of ``Loss weights" show that all settings with different scales of the loss weights $\lambda_{1}$ and $\lambda_{2}$ outperform the baseline, demonstrating that our method is insensitive to the loss weight. Moreover, our current setting ($\lambda_{1}=0.1$ and $\lambda_{2}=0.1$) achieves the best performance among other loss weight settings, demonstrating that selecting the proper loss weight to balance the scale of different loss items is helpful for improving the performance.
5) The results of the temperature parameter $\tau$ show that our method is not sensitive to the selection of the temperature parameter and 
the performance is best in most metrics when the temperature parameter is set to 0.5.

\noindent\textbf{Comparison of different LLMs.} 
We further conduct quantitative experiments to analyze how different LLMs impact the navigation performance. Specifically, we use different LLMs for navigational sequential landmark extraction,  including ChatGPT (used by our CONSOLE), GPT-3~\cite{brown2020language} (the predecessor of ChatGPT), Vicuna~\cite{vicuna} (a recent strong open source alternative to ChatGPT which has two sizes——7B and 13B), and Spacy~\cite{spacy} (a NLP library), and feed the extracted landmarks into the consequent modules. We access the APIs of ChatGPT and Vicuna~\cite{vicuna} in May, 2023. We access the API of  GPT-3~\cite{brown2020language} in Dec, 2022. The results are given in Table~\ref{tab:LLMs}. From Table~\ref{tab:LLMs} we can observe that: 1) ChatGPT outperforms other competitors in most metrics; 2) GPT-3, Vicuna13B, and Vicuna7B are comparable to ChatGPT in some metrics; 3) Vicuna13B outperforms Vicuna7B especially in SPL; 4) Spacy performs the worst among all competitors. These results show that more powerful LLMs can provide more accurate knowledge for navigation.

\begin{figure*}[t]
\begin{centering}
\includegraphics[width=0.98\linewidth]{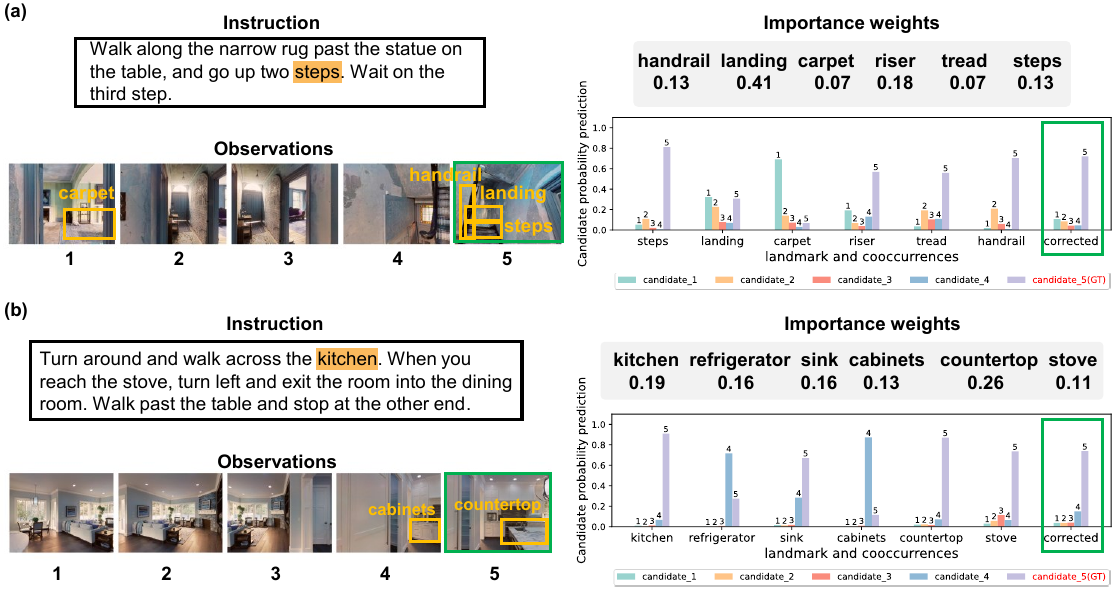}
\par\end{centering}
\caption{Visualization examples of action decision. The ground-truth (GT) action and the corrected landmark prediction are denoted in the green boxes.
}
\label{fig:visualiation}
\vspace{-0.2cm}
\end{figure*}

\begin{figure*}[t]
\begin{centering}
\includegraphics[width=0.98\linewidth]{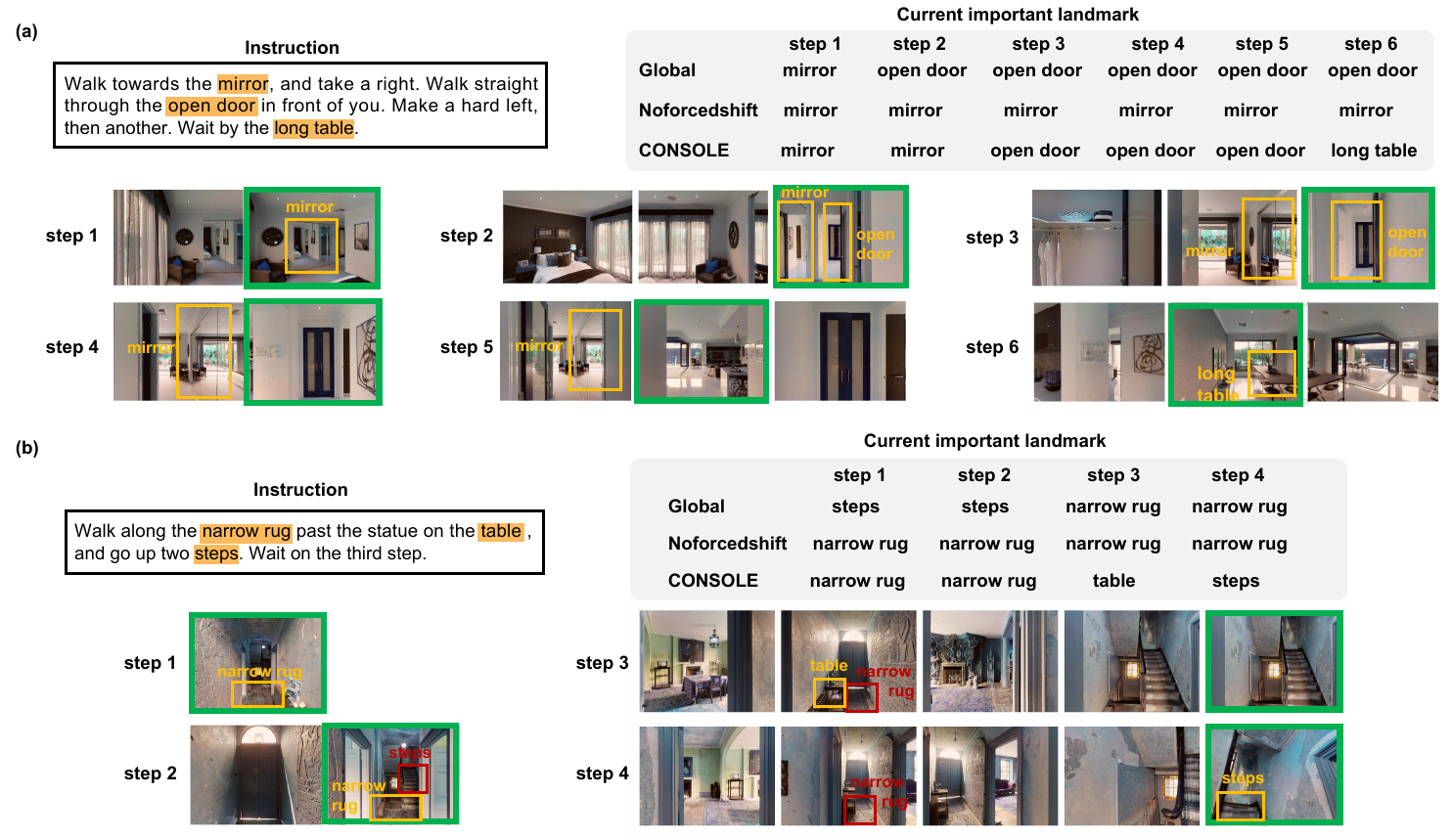}
\par\end{centering}
\caption{Visualization example of landmark shifting. The ground-truth action (GT) and the mentioned landmark are annotated in the green boxes and orange boxes, respectively. 
}
\label{fig:visualiation_shift}
\vspace{-0.2cm}
\end{figure*}

\begin{figure*}
\begin{centering}
\includegraphics[width=0.98\linewidth]{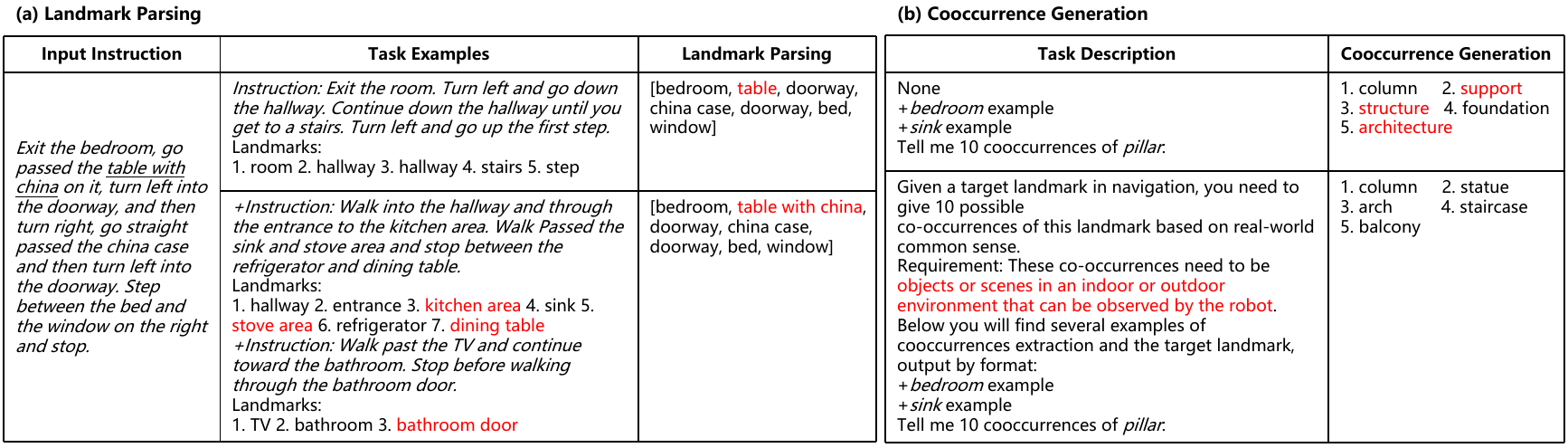}
\par\end{centering}
\caption{Visualization of comparison of different task examples and prompts.  
}
\label{fig:taskexample}
\vspace{-0.2cm}
\end{figure*}

\begin{figure*}[t]
\begin{centering}
\includegraphics[width=0.98\linewidth]{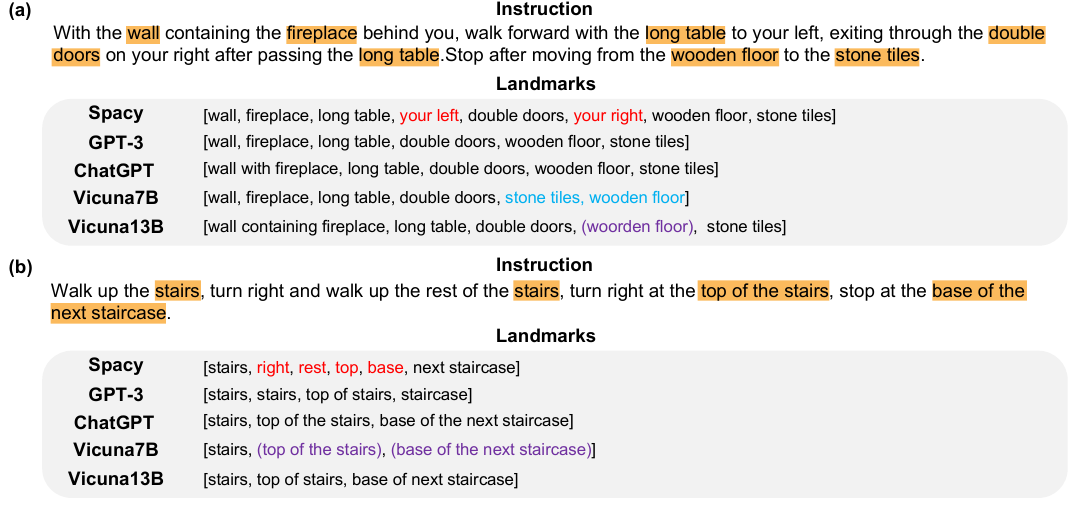}
\par\end{centering}
\caption{Comparison of navigational sequential landmark extraction among different LLMs. The correct landmarks are denoted in orange. For extraction results, noise is marked in red, landmarks in the wrong order are marked in blue, and missing landmarks are marked in purple with brackets.
}
\label{fig:visualiation_llm}
\vspace{-0.2cm}
\end{figure*}

\begin{figure*}
\begin{centering}
\includegraphics[width=0.98\linewidth]{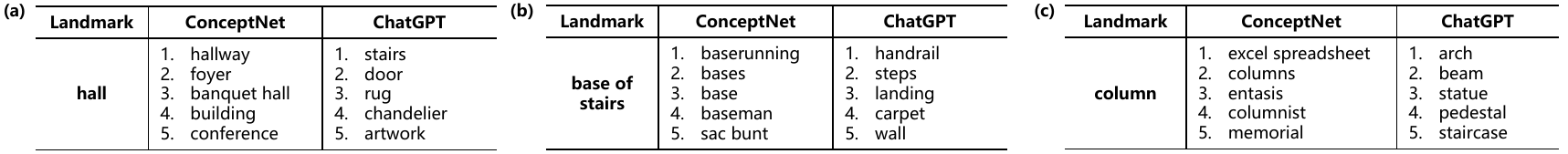}
\par\end{centering}
\caption{Comparison results of obtaining cooccurrences between ChatGPT and ConceptNet.  
}
\label{fig:conceptnet}
\vspace{-0.2cm}
\end{figure*}

\subsection{Visualization}
In this subsection, we present rich visualization results to deeply analyze the effects of different modules in CONSOLE, including the learnable cooccurrence scoring module, the landmark shifting module, the customized designs of prompts, {\it etc}.

\noindent\textbf{Action Decision.}
Fig.~\ref{fig:visualiation} gives some visualization examples of action decision, where we can find that CONSOLE effectively corrects the landmark cooccurrence priors for facilitating action prediction.
For example, in  Fig.~\ref{fig:visualiation}(a), the importance weight of  ``carpet'' which appears in  ``candidate\_1'' but does not appear in ``candidate\_5 (GT)'' is significantly low. And the importance weights of ``landing'', ``steps'', and ``handrail' which appear in ``candidate\_5 (GT)'' are relatively high. As a result, the landmark prediction is successfully corrected to indicate the GT action. 

\noindent\textbf{Landark shifting.}
Fig.~\ref{fig:visualiation_shift} shows that our landmark shifting module takes better advantage of the landmark navigation order for landmark discovery. 
It can effectively mitigate the stuck and disordered landmark prediction caused by scene complexity. For example, in Fig.~\ref{fig:visualiation_shift}(a) since the mirror appears in the observations during multiple steps, ``Noforcedshift'' continuously predicts ``mirror'' as the current important landmark. CONSOLE, however, predicts correct landmarks at different timesteps. 

\noindent\textbf{Impact of task examples and prompts.}
Fig.~\ref{fig:taskexample} gives the  visualization of landmark extraction and cooccurrence generation regarding different task examples and prompts. From Fig.~\ref{fig:taskexample} we can find that by providing task examples regarding diverse landmark descriptions, we can successfully improve the landmark extraction for LLM. For example, in Fig.~\ref{fig:taskexample}(a), when prompted with only one example that extracts landmarks of a single noun, ChatGPT only extracts {\it table}. After adding the two examples that extract noun-phrase landmarks, {\it table with china} can be successfully extracted. 

Moreover, we can also observe that adding the constraint in the prompt can effectively enable LLM to generate desired occurrences. For example, in Fig.~\ref{fig:taskexample}(b), ChatGPT generates abstract words such as {\it support} without the constraint. However, through adding the constraint, ChatGPT generates more reliable cooccurrences like {\it statue}.

\noindent\textbf{Navigational sequential landmark extraction.}
To verify the capability of existing LLMs for accurate landmark extraction, we give the visualization examples of landmark extraction comparison among different methods in Fig.~\ref{fig:visualiation_llm}. 
From Fig.~\ref{fig:visualiation_llm}, we can find that: (1) ChatGPT performs the best generally among all compared models; (2) Vicuna13B~\cite{vicuna} is comparable to ChatGPT while sometimes extracting wrong navigational order; (3) Vicuna7B~\cite{vicuna} also suffers from wrong navigational order (e.g., in Fig.~\ref{fig:visualiation_llm}(a), Vicuna7B extracts the ``stone tiles'' before extracting the ``wodden floor''); (4) GPT-3~\cite{brown2020language} is slightly inferior to ChatGPT for fine-grained landmark extraction (e.g., in Fig.~\ref{fig:visualiation_llm}(a), GPT-3 extracts the ``wall'' and the ``fireplace'' rather than extracting the ``wall containing fireplace'');  (5) Spacy~\cite{spacy} can only extract the textual order rather than the navigational order and extracts a lot of noise (e.g., in Fig.~\ref{fig:visualiation_llm}(b), Spacy extracts the ``right'', the ``rest'', and the ``base'').
These landmark extraction visualization results reveal the great potential of utilizing LLMs for assisting VLN. Moreover, it is also important to resort to more low-cost LLMs such as Vicuna~\cite{vicuna} for helping embodied agents in the future.

\noindent\textbf{Landmark cooccurrence extraction.}
To verify the superiority of LLM in landmark cooccurrence extraction, we present the comparison of obtaining landmark cooccurrences between ChatGPT and ConceptNet in Fig.~\ref{fig:conceptnet}. To get the cooccurrences of a target landmark through ConceptNet, we respectively request its related terms and nodes which are connected with relations of 'AtLocation' and 'LocatedNear'. From Fig.~\ref{fig:conceptnet} we can observe that LLM can generate visible cooccurrences as constrained in the prompt (Sec.~\ref{Landmark Cooccurrence Prior Generation}) which are informative and high-quality. ConceptNet, however, tends to generate synonyms and concepts rather than visible cooccurrences. For example, in Fig.~\ref{fig:conceptnet} (a), for the target {\it hall}, ChatGPT generates multiple cooccurrences of {\it hall} like {\it artwork} and {\it chandelier} to help recognize the location of the {\it hall}. However, ConceptNet retrieves {\it hallway} and {\it building} which are uninformative. 

Moreover, LLM can adapt to flexible inputs and process them robustly, whereas ConceptNet may generate poor results without  preprocessing. In Fig.~\ref{fig:conceptnet} (b), for the target {\it base of stairs} (landmark formed as a phrase), ChatGPT generates valid cooccurrences of stairs such as {\it handrail}. However, ConceptNet is unable to focus on the important part ({\it stairs}) and instead searches invalid related terms of {\it base}.


\section{Conclusion}
In this paper, we propose CONSOLE for VLN, which casts VLN as an open-world sequential landmark discovery problem by introducing a correctable landmark discovery framework based on two powerful large models. We harvest rich landmark cooccurrence commonsense from ChatGPT and employ CLIP for conducting landmark discovery. A learnable cooccurrence scoring module is constructed to correct the priors provided by ChatGPT according to actual observations.  Experimental results show that CONSOLE outperforms strong baselines consistently on R2R, REVERIE, R4R, and RxR. CONSOLE establishes new SOTA results on R2R and R4R under unseen scenarios. %

We believe that our work can provide a meaningful reference to the researchers in both the VLN and the embodied AI areas in how to effectively harvest and utilize the helpful knowledge inside large models for assisting robotic tasks, which is a very promising way to improve the performance of robots. In future work, we would like to resort to more low-cost large models for assisting embodied AI tasks, which is more practical in real-world applications. It is also worth studying to develop efficient in-domain pretraining paradigms to adapt large models to embodied AI tasks.

	
	%
\appendix
\label{Prompts with Task Examples}
\section*{Navigational Sequential Landmark Extraction Prompts}
\label{Landmark Extraction Task Example}
In this subsection, we present the prompt with task examples of landmark extraction. 
Specifically, we state in the prompt to 1) ask ChatGPT to extract navigational sequential landmarks rather than textual sequential landmarks, and 2) constrain ChatGPT not to generate abstract landmarks and landmarks not mentioned in the instruction. 
We use the same task examples for R2R, R4R and RxR since the instructions for them are all fine-grained.
For REVERIE which have high-level instructions, we pre-define 5 different task examples. Through being provided the customized prompt with diverse task examples, ChatGPT is able to extract navigational sequential landmarks with different kinds of descriptions, such as landmarks composed of several nouns and landmarks with adjectives. 

For a given instruction $I$, the whole prompt with task examples for the landmark extraction on R2R, R4R, and RxR are as follows:

\texttt{Given an instruction, you need to extract the landmarks in the instruction and sort them in the order in which they appear in the real navigation (not in the order they appear in the instruction). Landmarks must be the actual objects and scenes you see in the navigation, and do not include other abstract nouns such as ``left'' and ``right''.}

\texttt{Requirement 1: Extract all landmarks in the instruction.}

\texttt{Requirement 2: do not generate landmarks that are not in the instruction.}

\texttt{Below you will find several examples of landmark extraction and the instruction you need to complete the extraction, output by format:}

\texttt{Instruction: Exit the room. Turn left and go down the hallway. Continue down the hallway until you get to the stairs. Turn left and go up the first step.}

\texttt{Landmarks:}

\texttt{1. room; }

\texttt{2. hallway; }

\texttt{3. hallway; }

\texttt{4. stairs; }

\texttt{5. step. }

\texttt{Instruction: Walk into the hallway and through the entrance to the kitchen area. Walk Passed the sink and stove area and stop between the refrigerator and dining table.}

\texttt{Landmarks:}

\texttt{1. hallway; }

\texttt{2. entrance; }

\texttt{3. kitchen area; }

\texttt{4. sink; }

\texttt{5. stove area; }

\texttt{6. refrigerator; }

\texttt{7. dining table. }

\texttt{Instruction: Walk past the TV and continue toward the bathroom. Stop before walking through the bathroom door.}

\texttt{Landmarks:}

\texttt{1. TV; }

\texttt{2. bathroom; }

\texttt{3. bathroom door. }

\texttt{Instruction: Walk between the columns and make a sharp turn right. Walk down the steps and stop on the landing.}

\texttt{Landmarks:}

\texttt{1. columns; }

\texttt{2. steps; }

\texttt{3. landing. }

\texttt{Instruction: With the windows on your left, walk through the large room past the sitting areas. Go through the door left of the tapestry and enter a wood-paneled room with a circular table in the middle. Go up the stairs and stop on the sixth step from the bottom.}

\texttt{Landmarks:}

\texttt{1. windows; }

\texttt{2. large room; }

\texttt{3. sitting areas; }

\texttt{4. door; }

\texttt{5. tapestry; }

\texttt{6. wood-paneled room; }

\texttt{7. circular table; }

\texttt{8. stairs; }

\texttt{9. step. }

\texttt{Instruction: $I$}
          
\texttt{Landmarks:}

The task examples on REVERIE are shown below:

\texttt{Instruction: Go to the lounge on the first level and bring the trinket with the clock that's sitting on the fireplace.}

\texttt{Landmarks:}

\texttt{1. first level; }

\texttt{2. lounge; }

\texttt{3. fireplace; }

\texttt{4. clock; }

\texttt{5. trinket. }

\texttt{Instruction: Go to the staircase by entryway and touch the front of the banister of the staircase.}

\texttt{Landmarks:}

\texttt{1. entryway; }

\texttt{2. staircase; }

\texttt{3. staircase; }

\texttt{4. banister. }

\texttt{Instruction: Go to the bedroom with the fireplace and bring me the lowest hanging small picture on the right wall across from the bedside table with the lamp on it.}

\texttt{Landmarks:}

\texttt{1. bedroom; }

\texttt{2. fireplace; }

\texttt{3. bedside table; }

\texttt{4. lamp; }

\texttt{5. wall; }

\texttt{6. small picture. }

\texttt{Instruction: Go to the bedroom on level 2 to the right of the green bathroom and remove the white pillow closest to the bedroom door from the bed.}

\texttt{Landmarks:}

\texttt{1. level 2; }

\texttt{2. green bathroom; }

\texttt{3. bedroom; }

\texttt{4. bedroom door; }

\texttt{5. bed; }

\texttt{6. white pillow. }

\texttt{Instruction: Go to the first level bedroom adjacent to the hallway leading to the lounge and dust the sofa chair and place 2 more pillows on it.}

\texttt{Landmarks:}

\texttt{1. first level; }

\texttt{2. hallway; }

\texttt{3. lounge; }

\texttt{4. bedroom; }

\texttt{5. sofa chair; }

\texttt{6. pillows. }

\section*{Landmark Cooccurrence Generation Prompts}
\label{Co-occurrence Extraction Task Example}
In this subsection, we give the prompt with task examples of cooccurrence extraction. We state in the prompt to constrain ChatGPT to generate visible objects or scenes rather than abstract objects (e.g., wind) as cooccurrences. For the $k$-th landmark $U_{k}^{la}$ in the extracted landmark lists of instruction $I$, the whole prompt with task examples are presented below:

\texttt{Given a target landmark in navigation, you need to give 10 possible co-occurrences of this landmark based on real-world common sense.}

\texttt{Requirement: These co-occurrences need to be objects or scenes in an indoor or outdoor environment that can be observed by the robot.}


\texttt{Below you will find several examples of co-occurrences extraction and the target landmark, output by format:}

\texttt{Tell me 10 co-occurrences of bedroom:}

\texttt{1. bed; }

\texttt{2. door; }

\texttt{3. window; }

\texttt{4. mirror; }

\texttt{5. closet; }

\texttt{6. rug; }

\texttt{7. curtains; }

\texttt{8. walls; }

\texttt{9. ceiling; }

\texttt{10. floor. }

\texttt{Tell me 10 co-occurrences of sink:}

\texttt{1. water; }

\texttt{2. faucet; }

\texttt{3. basin; }

\texttt{4. counter; }

\texttt{5. tile; }

\texttt{6. porcelain; }

\texttt{7. chrome; }

\texttt{8. soap; }

\texttt{9. towel; }

\texttt{10. mirror. }

\texttt{Tell me 10 co-occurrences of  $U_{k}^{la}$:}

We empirically find that 2 task examples for the cooccurrence generation are enough in CONSOLE.
	\section*{Acknowledgments}

	This work was supported in part by National Science and Technology Major Project (2020AAA0109704),  Guangdong Outstanding Youth Fund (Grant No. 2021B1515020061), Mobility Grant Award under Grant No.  M-0461,  Shenzhen Science and Technology Program (Grant No. GJHZ20220913142600001), Nansha Key RD Program under Grant No.2022ZD014, CAAI-Huawei MindSpore Open Fund. We thank MindSpore for the partial support of this work, which is a new deep learning computing framwork\footnote{https://www.mindspore.cn/}.

	\ifCLASSOPTIONcaptionsoff
	\newpage
	\fi

	
	
	%
		
		
\vspace{-0.2cm}
\bibliographystyle{IEEEtran}
\bibliography{IEEEabrv,egbib}	
	
	%

\begin{IEEEbiography}[{\includegraphics[width=1in,height=1.25in,clip,keepaspectratio]{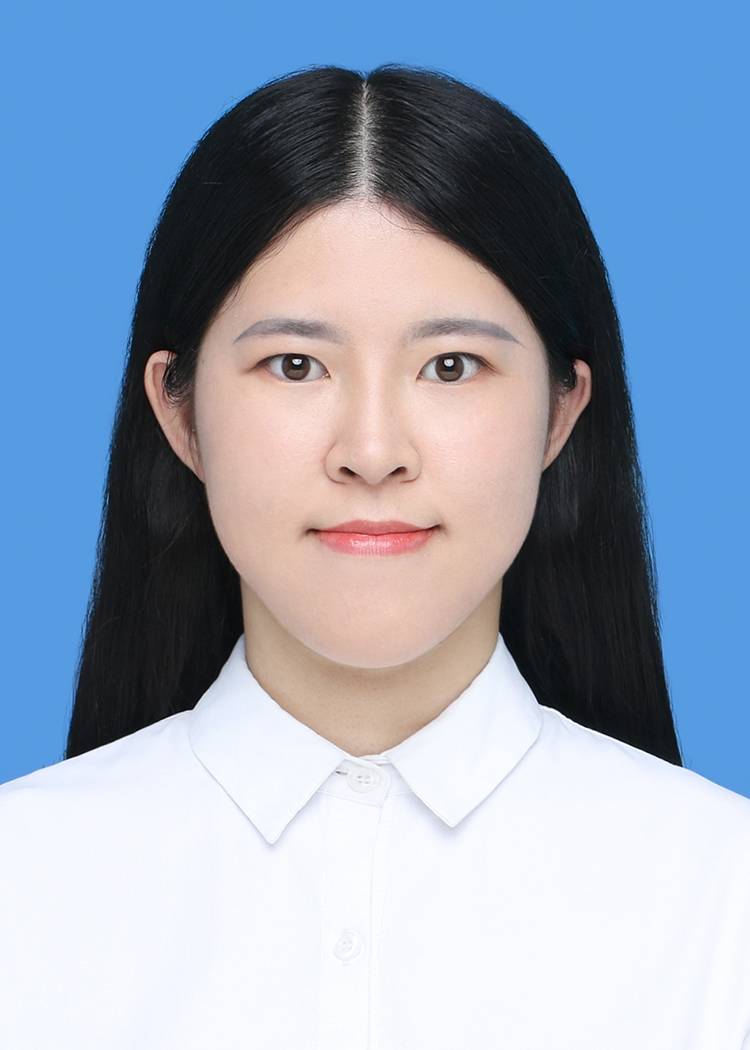}}]{Bingqian Lin} received the B.E. and the M.E.
degree in Computer Science from University of
Electronic Science and Technology of China and
Xiamen University, in 2016 and 2019, respectively.
She is currently working toward the D.Eng in the
school of intelligent systems engineering of Sun Yat-sen University. Her research interests include multi-view clustering, image processing and vision-and-language understanding.
\end{IEEEbiography}

\begin{IEEEbiography}[{\includegraphics[width=1in,height=1.25in,clip,keepaspectratio]{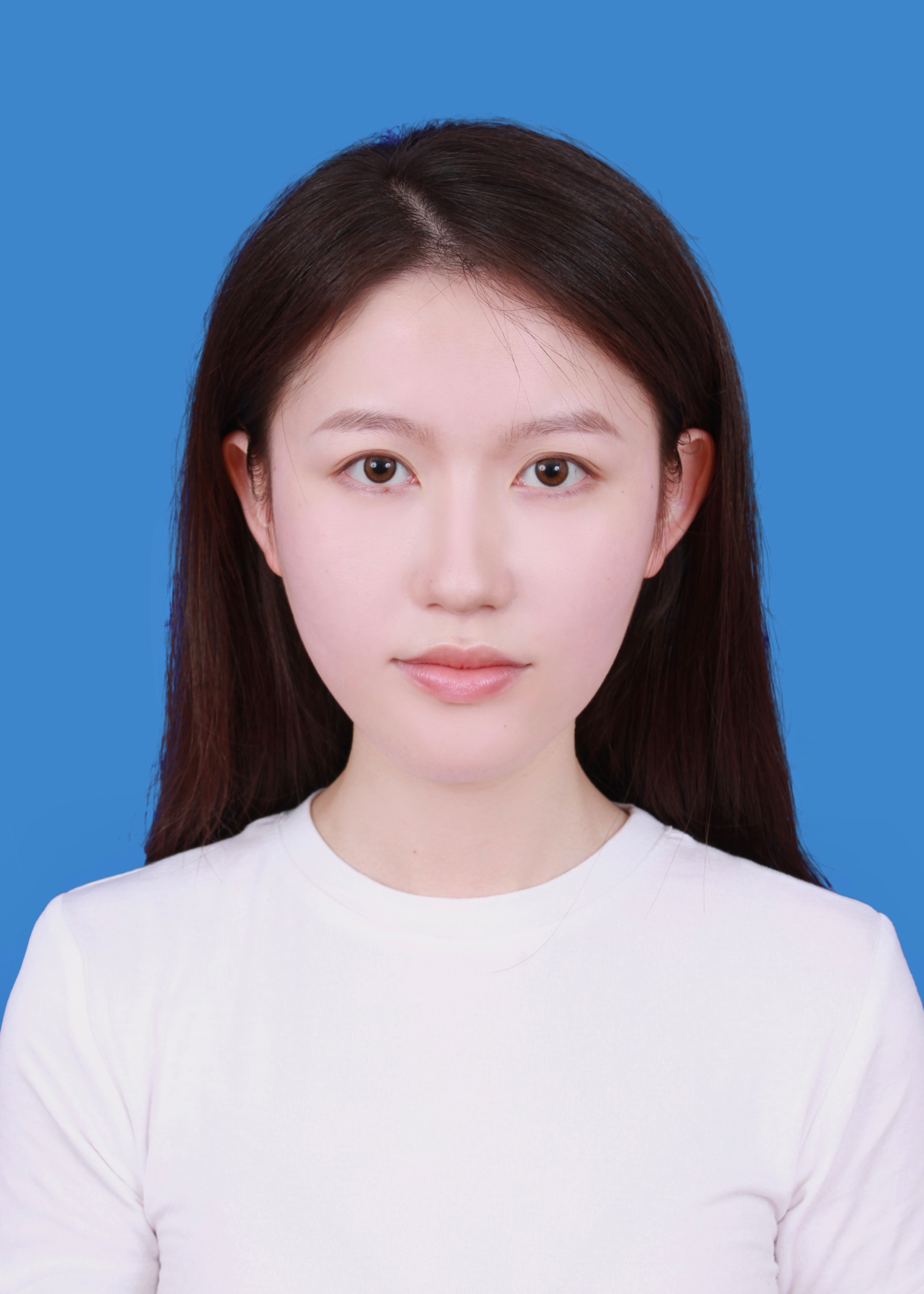}}]{Yunshuang Nie} received the B.E. degree in Sun Yat-sen University, Shenzhen, China, in 2023. She is currently working toward the M.E. in the school of intelligent systems engineering of Sun Yat-sen University. Her current research interest is vision-and-language understanding.
\end{IEEEbiography}

\begin{IEEEbiography}[{\includegraphics[width=1in,height=1.25in,clip,keepaspectratio]{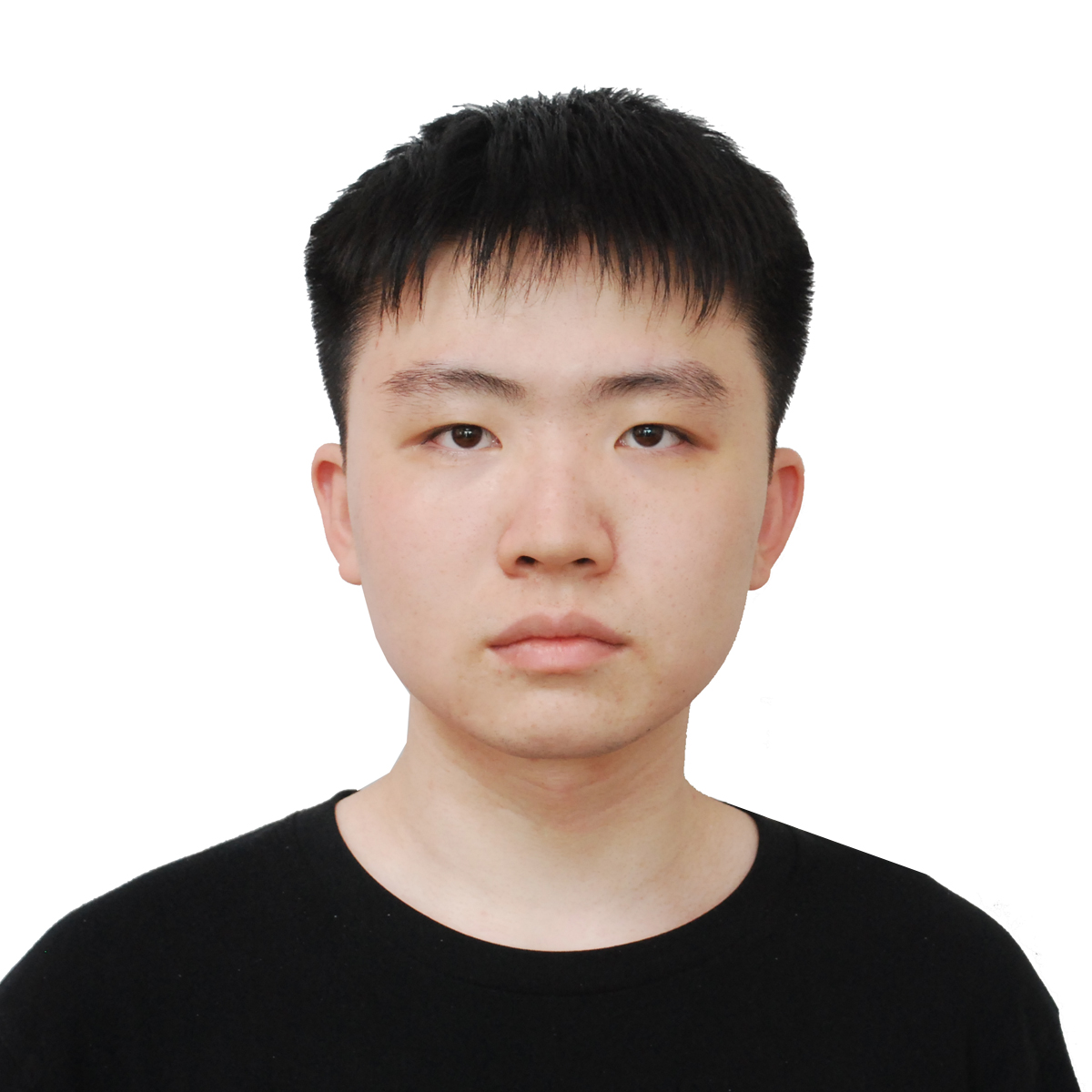}}]{Ziming Wei} is currently an undergraduate in the
school of intelligent systems engineering of Sun Yat-sen University. His current research interests include vision-and-language understanding, multimodality and embodied agent.
\end{IEEEbiography}

\begin{IEEEbiography}[{\includegraphics[width=1in,height=1.25in,clip,keepaspectratio]{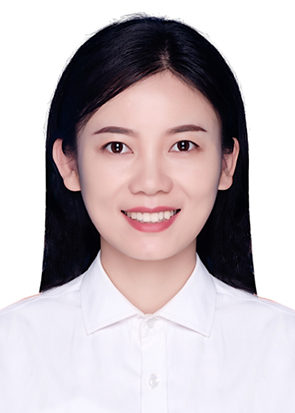}}]{Yi Zhu} received the B.S. degree in software engineering from Sun Yat-sen University, Guangzhou, China, in 2013. Since 2015, she has been a Ph.D student in computer science at the School of Electronic, Electrical, and Communication Engineering, University of Chinese Academy of Sciences, Beijing, China. Her current research interests include object recognition, scene understanding, weakly supervised learning, and visual reasoning.
\end{IEEEbiography}
 
\begin{IEEEbiography}
[{\includegraphics[width=1in,height=1.25in, clip,keepaspectratio]{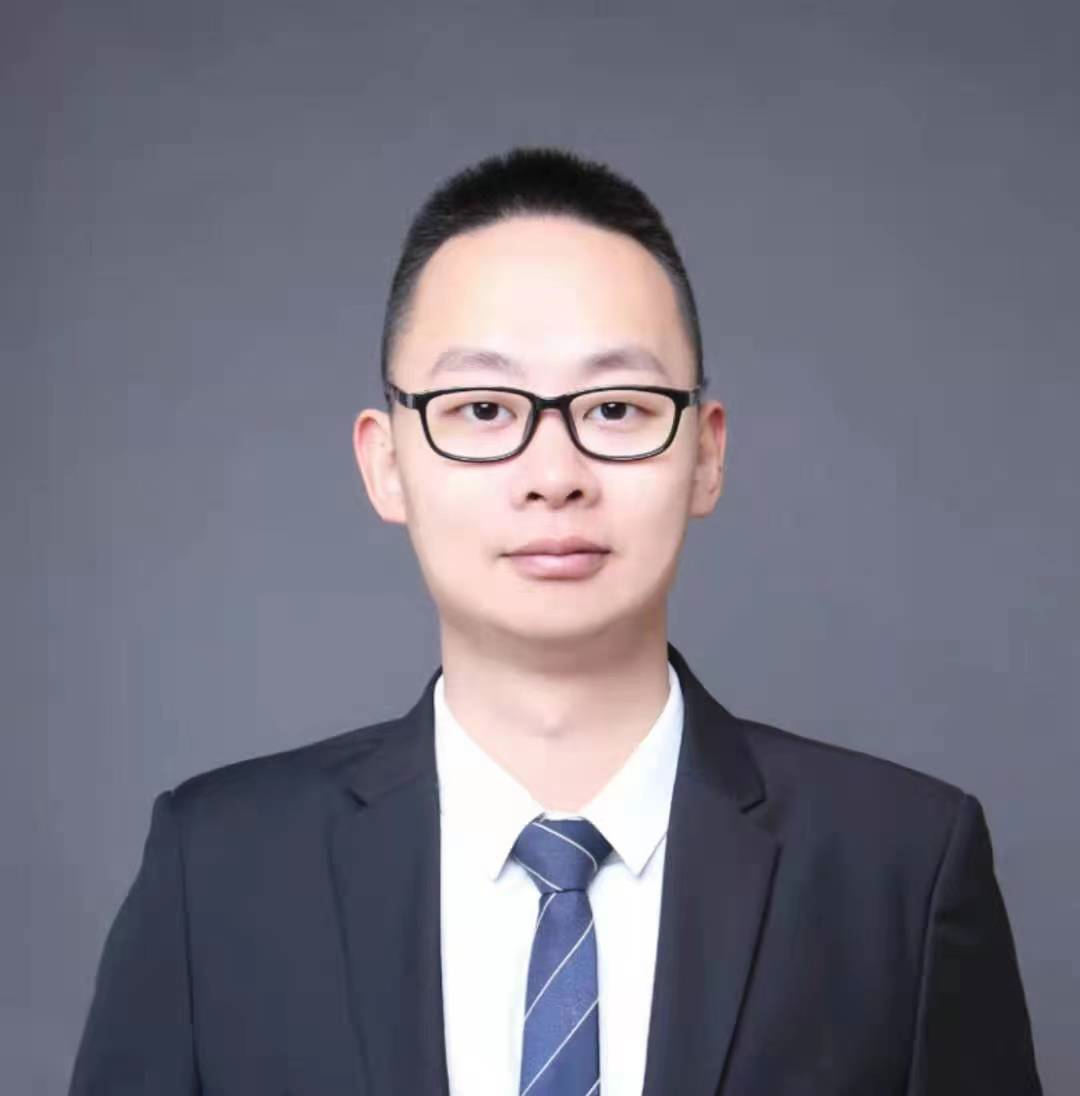}}]{Hang Xu} is currently a senior researcher in Huawei Noah Ark Lab. He received his BSc in Fudan University and Ph.D in Hong Kong University in Statistics. His research Interest includes multi-modality learning, machine Learning, object detection, and AutoML. He has published 70+ papers in Top AI conferences: NeurIPS, CVPR, ICCV, AAAI and some statistics journals, e.g. CSDA, Statistical Computing.
\end{IEEEbiography}

\begin{IEEEbiography}
[{\includegraphics[width=1in,height=1.25in, clip,keepaspectratio]{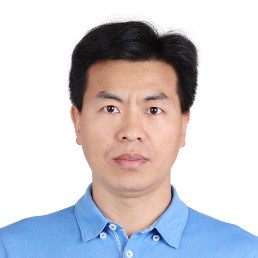}}]{Shikui Ma} received the M.S. degree in Northern Jiaotong University in 2003. He has been serving as an assistant vice president of Dataa Robotics company (https://www.dataarobotics.com/en) since 2015, where he is leading HARIX and AI R\&D team to consistently enhance their HARIX intelligent system, especially significantly improve the smart vision and motion capabilities of their robots via real-time digital twin, multi-modal fusion perception and advanced cognition, and deep reinforcement learning technologies. He has approximately 19 years of experience in communications technology industry, expertized in large-scale carrier-grade applications, especially competitive in Robotics, AI, Cloud, OSS and IPTV fields.
\end{IEEEbiography}

\begin{IEEEbiography}[{\includegraphics[width=1in,height=1.25in,clip,keepaspectratio]{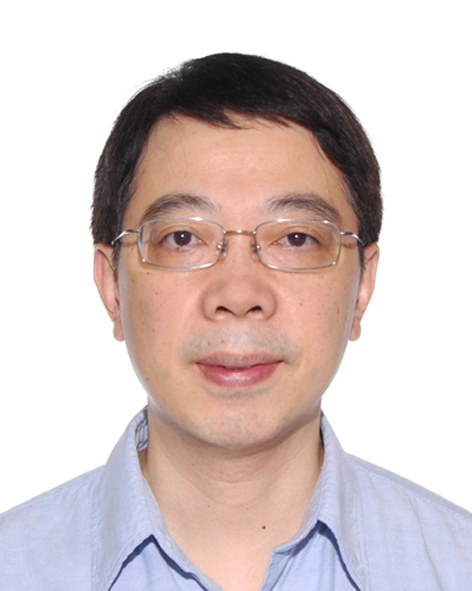}}]{Jianzhuang Liu} (Senior Member, IEEE) received the PhD degree in computer vision from The Chinese University of Hong Kong, in 1997. From 1998 to 2000, he was a research fellow with Nanyang Technological University, Singapore. From 2000 to 2012, he was a post-doctoral fellow, an assistant professor, and an adjunct associate professor with The Chinese University of Hong Kong, Hong Kong. In 2011, he joined the Shenzhen Institute of Advanced Technology, University of Chinese Academy of Sciences, Shenzhen, China, as a professor. He was a principal researcher in Huawei Company from 2012 to 2023. He has authored more than 200 papers in the areas of computer vision, image processing, deep learning, and AIGC.
\end{IEEEbiography}

\begin{IEEEbiography}[{\includegraphics[width=1in,height=1.25in,clip,keepaspectratio]{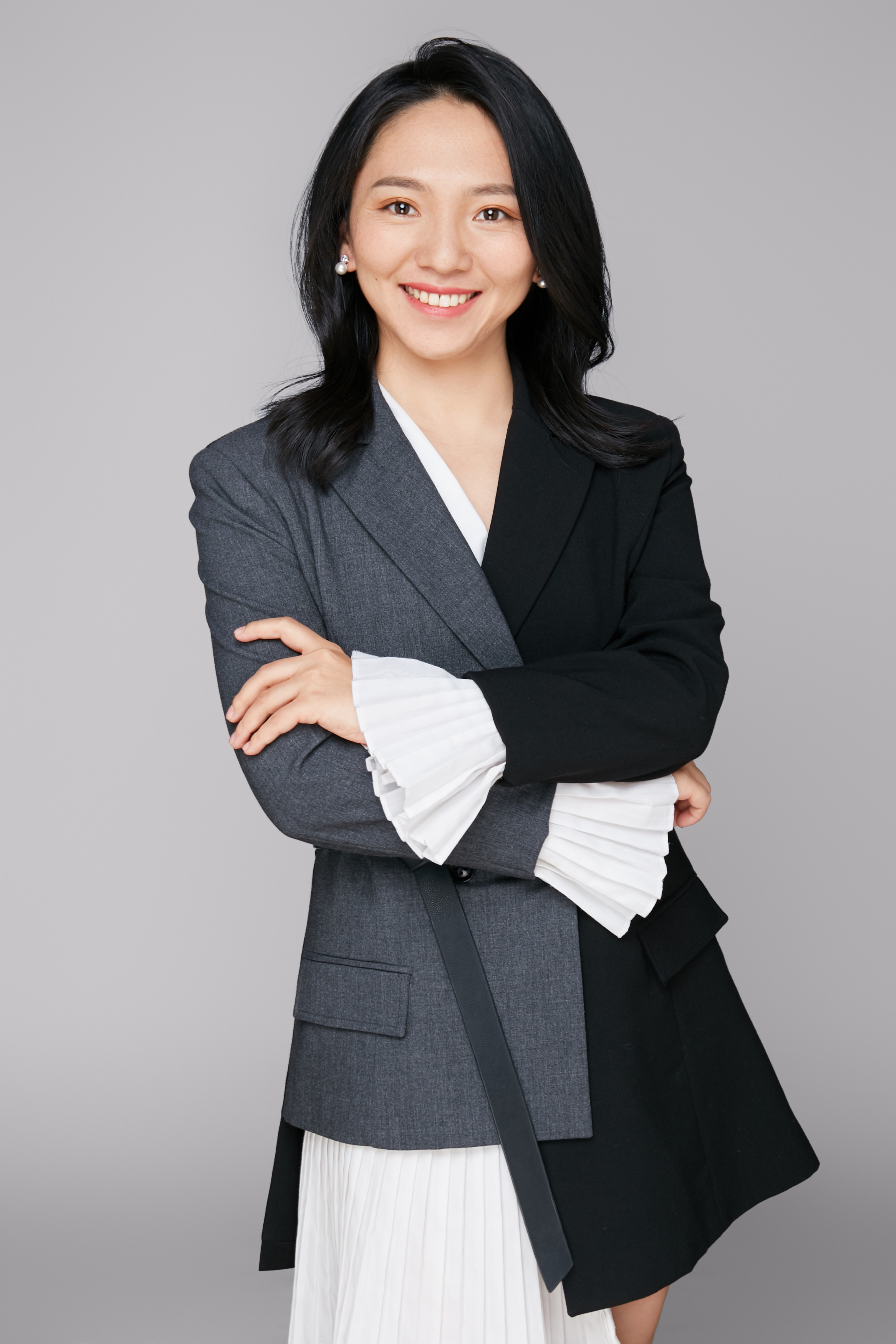}}]{Xiaodan Liang} is currently an Associate Professor at Sun Yat-sen University. She was a postdoc researcher in the machine learning department at Carnegie Mellon University, working with Prof. Eric Xing, from 2016 to 2018. She received her PhD degree from Sun Yat-sen University in 2016, advised by Liang Lin. She has published several cutting-edge projects on human-related analysis, including human parsing, pedestrian detection and instance segmentation, 2D/3D human pose estimation, and activity recognition.
\end{IEEEbiography}

	
	
	
	
	
	

\end{document}